\documentclass[10pt,twocolumn,letterpaper]{article}

\usepackage[pagenumbers]{cvpr} 

\usepackage{etoc}
\newcommand{\oursfull}{Chain-of-Visual-Thought\xspace}
\newcommand{\ours}{\textsc{CoVT}\xspace}

\newcommand{\relpct}[2]{%
  \begingroup
  \pgfmathsetmacro{\base}{#1}%
  \pgfmathsetmacro{\val}{#2}%
  \pgfmathsetmacro{\pctraw}{\val-\base}%
  \pgfmathtruncatemacro{\sgn}{sign(\pctraw)}%
  \pgfmathsetmacro{\absround}{round(10*abs(\pctraw))/10}%
  \pgfmathtruncatemacro{\iszero}{abs(\pctraw) < 0.05 ? 1 : 0}%

  \def\colorname{gray}\def\prefix{+}%
  \ifnum\sgn=1 \def\colorname{green!60!black}\def\prefix{+}\fi
  \ifnum\sgn=-1 \def\colorname{red!70!black}\def\prefix{-}\fi
  \ifnum\iszero=1 \def\colorname{gray}\def\prefix{+}\fi

  \makebox[0pt][l]{%
    \smash[b]{\raisebox{-0.2ex}{%
      \scriptsize\textcolor{\colorname}{\,(\prefix\pgfmathprintnumber[fixed,precision=1]{\absround})}%
    }}%
  }%
  \endgroup
}

\newcommand{\val}[2]{#2\relpct{#1}{#2}}

\usepackage{amssymb} 
\usepackage[table]{xcolor}
\usepackage{tikz}
\usepackage{float}
\usepackage{cuted}
\usepackage{booktabs}
\usepackage{makecell}
\usepackage{multirow}
\usepackage[table]{xcolor}
\usepackage{amssymb}
\usepackage{xcolor}

\newcommand{\xmark}{\ding{55}}
\usepackage{pifont}
\definecolor{cvprblue}{rgb}{0.21,0.49,0.74}
\definecolor{impgreen}{rgb}{0.1, 0.5, 0.1}
\definecolor{ForestGreen}{rgb}{0.13, 0.55, 0.13}
\usepackage[pagebackref,breaklinks,colorlinks,allcolors=cvprblue]{hyperref}

\def\paperID{5603} 
\def\confName{CVPR}
\def\confYear{2026}

\title{Chain-of-Visual-Thought: \\Teaching VLMs to See and Think Better with Continuous Visual Tokens}

\author{
\textbf{Yiming Qin}$^{1,4}$\quad
\textbf{Bomin Wei}$^{2}$\quad
\textbf{Jiaxin Ge}$^{1}$\quad
\textbf{Konstantinos Kallidromitis}$^{3}$ \\
\textbf{Stephanie Fu}$^{1}$\quad
\textbf{Trevor Darrell}$^{1}$\quad
\textbf{XuDong Wang}$^{1,4}$\footnotemark[2]\\[0.2cm]
$^{1}$UC Berkeley \quad
$^{2}$UCLA \quad
$^{3}$Panasonic AI Research \quad
$^{4}$Duke University
}

\newcommand{\ghlink}{\href{https://github.com/Wakals/CoVT}{https://github.com/Wakals/CoVT}}
\newcommand{\projpagelink}{\href{https://wakalsprojectpage.github.io/covt-website/}{https://wakalsprojectpage.github.io/covt-website/}}

\begin{document}
\twocolumn[{%
\maketitle
\begin{center}
\vspace{-15pt}
\etocdepthtag.toc{mtchapter}\includegraphics[width=\textwidth]{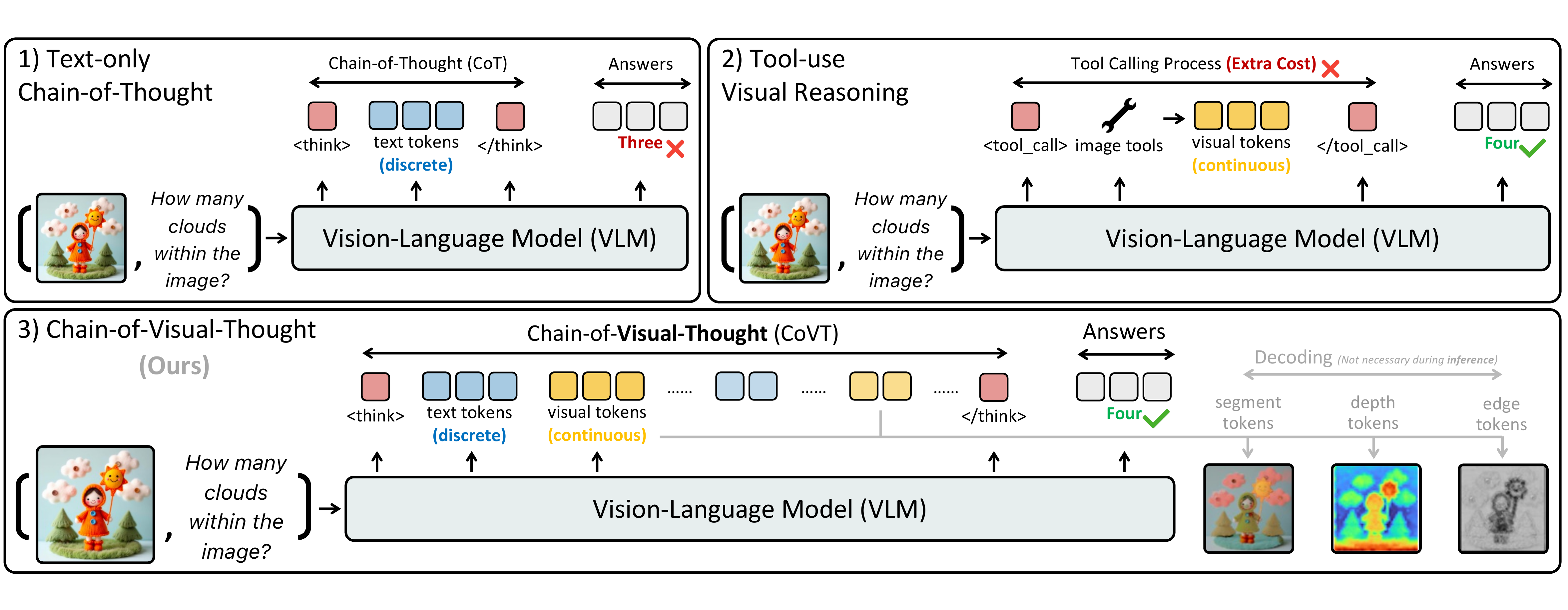}
\captionof{figure}{
Traditional VLM reasoning is restricted to the discrete language space (\textit{Text-only CoT}) or relies on costly external tool calls to obtain visual signals (\textit{Tool-use reasoning}).
\textbf{In contrast, \ours forms a visual thought chain that enables VLMs to reason in continuous visual space.}
By introducing \emph{continuous visual tokens} that encode perceptual cues (\eg, segmentation, depth, instance, and edge structure), 
\ours composes \emph{chains of textual and visual thoughts} that link semantic reasoning with perceptual grounding. 
These visual ``thought chains'' bridge language and vision, enabling fine-grained understanding, spatial precision, and geometric awareness beyond the reach of text-based reasoning. During inference, these visual tokens can be decoded into different types of visual information when interpretation is desired.}
\label{fig:teaser}
\end{center}
}]

\renewcommand{\thefootnote}{\fnsymbol{footnote}}
\footnotetext[2]{Corresponding authors.}
\renewcommand{\thefootnote}{\arabic{footnote}}

\begin{abstract}
    %
    
    Vision–Language Models (VLMs) excel at reasoning in linguistic space but struggle with perceptual understanding that requires dense visual perception, \eg, spatial reasoning and geometric awareness.
    This limitation stems from the fact that current VLMs have limited mechanisms to capture dense visual information across spatial dimensions.
    %
    We introduce \textbf{Chain-of-Visual-Thought} (\ours), a framework that enables VLMs to reason not only in words but also through \emph{continuous visual tokens}—compact latent representations that encode rich perceptual cues. Within a small budget of roughly 20 tokens, \ours distills knowledge from lightweight vision experts capturing complementary properties such as 2D appearance, 3D geometry, spatial layout, and edge structure.
    During training, the VLM with \ours
    autoregressively predicts these visual tokens to reconstruct dense supervision signals (\eg, depth, segmentation, edges, and DINO features).
    At inference, the model reasons directly in the continuous visual token space, preserving efficiency while optionally decoding dense predictions for interpretability.
    Evaluated across more than ten diverse perception benchmarks, including CV-Bench, MMVP, RealWorldQA, MMStar, WorldMedQA, and HRBench, integrating \ours into strong VLMs such as Qwen2.5-VL and LLaVA consistently improves performance by 3\% to 16\% and demonstrates that compact continuous visual thinking enables more precise, grounded, and interpretable multimodal intelligence.
    Our code is available at \ghlink.
\end{abstract}    
\section{Introduction}
\label{sec:intro}


\begin{figure*}[th]
    \centering
    \includegraphics[width=1.0\linewidth]{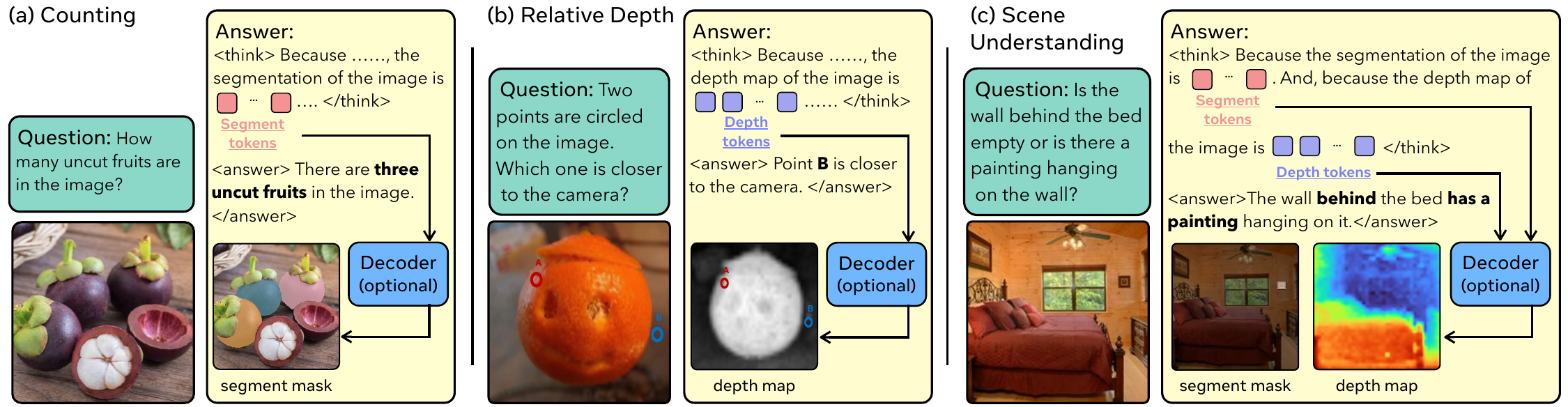}
    \caption{
    \textbf{Continuous visual thinking with \ours.}
    \ours introduces compact, continuous visual tokens that encode fine-grained perceptual cues, such as object localization, spatial structure, and scene semantics, directly into VLM reasoning. 
    These tokens ground multimodal reasoning in visual space, enabling the model to capture fine-grained relationships across vision-centric tasks (\eg, counting, depth ordering, and scene understanding) without relying on external tools. 
    They can also be decoded into dense predictions, offering human-interpretable visualizations of the model's reasoning process.
    }
    \label{fig:teaser_demo}
\end{figure*}

Vision–Language Models (VLMs)~\cite{Qwen2.5-VL,Liu2023ImprovedBWllava,team2023gemini,dubey2024llama,Zhai2023SigmoidLFSigLIP,achiam2023gpt,zhu2025internvl3,team2025gemma,xie2025reconstruction,deng2025emerging} have become the cornerstone of modern multimodal intelligence, achieving remarkable progress in understanding and reasoning across text and vision. 
By projecting visual input into a language-centric token space, VLMs inherit the strong compositional and logical reasoning capabilities of large language models (LLMs), enabling unified multimodal interaction through natural language. 
Recent advances in text-based Chain-of-Thought (CoT) reasoning~\cite{wei2022chain} further extend this paradigm, showing that structured intermediate reasoning steps can significantly enhance performance on tasks involving logic, mathematics, and knowledge grounding. 
However, despite these successes, such reasoning remains fundamentally \emph{language-bound}. 

When continuous visual information is projected into discrete text space, \textit{rich perceptual cues, \eg, boundaries, layout, depth, and geometry, are lost or poorly represented.} 
Yet these are precisely the fine-grained signals that humans rely on when reasoning about the visual world. 
Consequently, current VLMs often struggle with perception-intensive tasks such as counting, spatial correspondence, or relative depth estimation, even when equipped with powerful vision encoders~\citep{fu2025hiddenplainsightvlms, rahmanzadehgervi2025visionlanguagemodelsblind,wang2025visual}, as shown in \cref{fig:teaser}.
Moreover, by forcing vision reasoning through a discrete text bottleneck, the model must verbalize continuous spatial and geometric relations. As a result, text-only CoT can misdirect and even \emph{degrade} visual reasoning performance,
as shown by Qwen3-VL-Thinking~\citep{Yang2025Qwen3TR, Qwen2.5-VL}, which performs over 5\% worse than Qwen3-VL-Instruct with language CoT on spatial understanding benchmarks such as V\*~\cite{wu2024v}, HRBench8k~\cite{wang2025divide}, and VSI-Bench~\cite{yang2025thinking}. 
This exposes a fundamental limitation: \textit{\textbf{visual information is inherently continuous and high-dimensional, yet existing models reason over it using symbolic language tokens that lack the fidelity of complex perceptual reasoning.}}

A natural solution is to augment VLMs with external vision tools~\cite{gupta2023visual, suris2023vipergpt}, leveraging pre-built specialized models to recover fine-grained perception. 
While this approach can partially restore spatial and geometric information, it also introduces significant drawbacks: perception is delegated to external tools, and outcomes are bounded by them. It also introduces higher GPU cost.
Another solution is generating or cropping images in the thinking process. However, these solutions still project the images into the text space, losing the dense visual information.
These limitations motivate a central question:  
\textit{\textbf{Can VLMs learn to reason the way humans do, by thinking visually rather than translating everything into words?}}  
More concretely, can we inject fine-grained visual signals directly into a VLM's reasoning process, allowing it to ``see'' and ``think'' simultaneously while remaining efficient and self-contained?  
Yes! We propose \textbf{C}hain-\textbf{o}f-\textbf{V}isual-\textbf{T}houghts \textbf{(\ours)}. 

\ours enables reasoning over rich perceptual cues by grounding VLMs in continuous visual token space. 
Each group of visual tokens corresponds to a lightweight perceptual expert (\eg, segmentation, depth, edge detection, or self-supervised representation learning) that encodes specific visual features. 
During training, the VLM is asked to \emph{predict} these continuous visual tokens within its reasoning chain, 
compressing rich perceptual information into a compact latent space. 
These latent tokens are then decoded by task-specific lightweight decoders to reconstruct the corresponding expert targets (\eg, segmentation masks, dense depth maps, edge maps, or DINO features). 
We backpropagate the reconstruction and distillation losses through the continuous tokens, aligning the model's internal latent representations with expert guidance. 
This process allows \ours to internalize fine-grained perceptual knowledge directly into its token space, enabling grounded reasoning without explicit visual maps or external tool calls, as shown in \cref{fig:teaser}.

More specifically, we highlight different aspects of fine-grained visual reasoning.
We integrate both task-oriented experts (\eg, SAM~\citep{Kirillov2023SegmentASAM}, DepthAnything v2~\citep{Yang2024DepthAVDepthAnything}, PIDINet~\citep{Su2021PixelDNPIDINet}) and representation-based experts (\eg, DINO~\citep{Caron2021EmergingPIDINO}, contrastive encoders), with alignment strategies tailored to each: task-oriented signals are aligned at the prompt level, while representation-based signals are aligned in feature space.
Training proceeds through four stages, \textit{including comprehension, generation, reasoning, and efficient reasoning}, gradually teaching the model to reason effectively with visual thoughts.

At inference, the model forms \emph{chains of visual thoughts}, reasoning across modalities to produce answers that are both semantically coherent and perceptually grounded. 
This self-contained, differentiable process enables VLMs to ``think'' directly in continuous visual space, thereby providing a more faithful bridge between internal reasoning and perceptual understanding. 
Moreover, this design supports interpretable multimodal intelligence, allowing users to visualize the model’s visual thinking process when desired, as shown in \cref{fig:teaser_demo}.
If visualization is not required, \ours can operate solely on the continuous visual tokens without decoding them into dense predictions, thus maintaining efficiency.

Evaluated across diverse perception benchmark, \ours consistently improves fine-grained visual reasoning, outperforming strong VLM baselines on vision-centric tasks while maintaining competitive performance on general (non-vision-centric) benchmarks. 
For example, \ours achieves a 5.5\% overall gain on CV-Bench~\citep{Tong2024Cambrian1AFCVBench}, delivering a substantial 14.0\% improvement on its depth sub-task, and 4.5\% overall gain on HRBench~\citep{Wang2024DivideCAHRBench}.
In addition, \ours offers flexible interpretability: the continuous visual tokens can be decoded into human-readable dense predictions, providing a window into the model’s underlying visual reasoning process when desired. 
Together, these results demonstrate that compact continuous visual thinking enables more precise, grounded, and interpretable multimodal intelligence. 

\noindent \textbf{\textit{The main elements of our contribution are as follows:}}
\begin{itemize}[leftmargin=*, topsep=0pt]
\item We propose {Chain-of-Visual-Thought}, a framework that equips VLMs with the ability to reason through \emph{continuous visual tokens}, compact perceptual representations that serve as the building blocks for multimodal thinking.
\item We develop tailored alignment strategies and a training pipeline (\textit{comprehension, generation, reasoning, and efficient reasoning}) that enable VLMs to learn, interpret, and reason effectively within continuous visual space.
\item We demonstrate consistent performance gains across diverse benchmarks, showing that continuous visual tokens enhance both perceptual grounding and interpretability.
\end{itemize}

\begin{figure*}[t]
    \centering
    \includegraphics[width=1.0\linewidth]{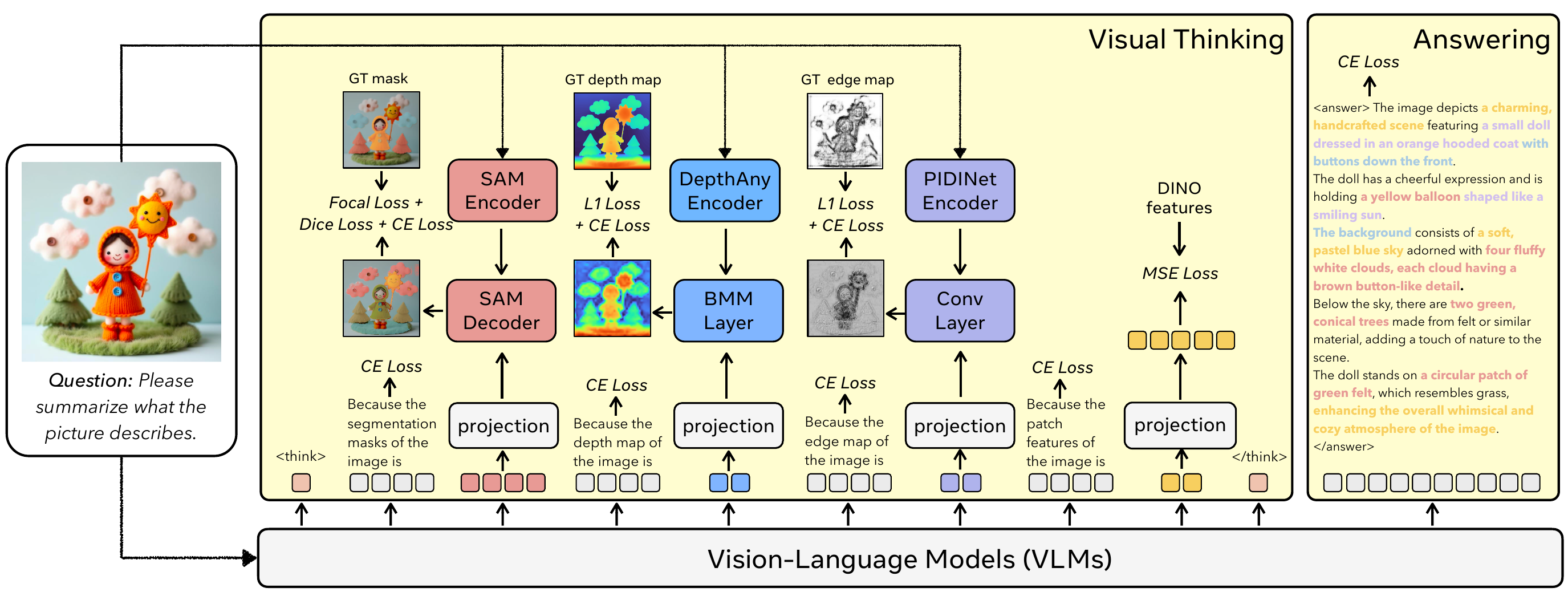}
    \caption{\textbf{The training pipeline of \ours.} \ours first generates the thinking process, containing visual thinking tokens, and then leverages these visual thoughts to condition next token prediction and reason the final answer.
    To endow these tokens with perceptual meaning, we align them with lightweight vision experts (\textit{e.g.}, SAM, DepthAnything, PIDINet, DINO) on their respective tasks during training.
    Specifically: SAM uses 8 visual tokens as mask prompts; DepthAnything uses 4 tokens to reconstruct depth; PIDINet uses 4 tokens to reconstruct edges; and DINO uses 4 tokens to match patch-level features. 
    The VLM is finetuned with LoRA and all the projection layers are trainable.
    \textit{\textbf{Note: During inference, dense predictions are decoded only when interpretability is desired; otherwise, reasoning occurs entirely in the latent visual space.}}
    }
    \label{fig:method-framework}
\end{figure*}

\section{Related Work}
\noindent \textbf{Tool-Augmented Reasoning}
Equipping VLMs with external tools enables them to use specialized vision models for targeted visual tasks~\citep{Lu2023ChameleonPC, Wu2023VisualCT, Yin2025ToolVQAAD, suris2023vipergpt, gupta2023visual}. 
While this improves performance, it also introduces computational overhead. 
Moreover, tool usage is inherently constrained, as the final performance is bounded by the ability of each tool instead of the reasoning process itself. 
In this work, we consider self-contained visual reasoning, which conducts reasoning flexibly and does not rely on external vision tools.

\noindent \textbf{Text Space Reasoning}
Text space reasoning methods, such as Chain-of-Thought ~\citep{Kojima2022LargeLMStep}, have achieved big success in language reasoning~\citep{wei2022chain, wang2024chainofthoughtreasoningprompting, madaan2023selfrefineiterativerefinementselffeedback, wang2023selfconsistencyimproveschainthought}, solving problems like math, science, and logical reasoning. The strong performance of LLMs with CoT capabilities has led to their broad adoption and success in models such as DeepSeek-R1~\citep{deepseekai2025deepseekr1incentivizingreasoningcapability}.
With the success of CoT, many works have extended reasoning to the visual modality.
A straightforward approach is to generate dense captions and reason in language space~\citep{lu2022learnexplainmultimodalreasoning}, but this process is inherently lossy.


We compare \ours with recent multimodal reasoning paradigms in Tab.~\ref{tab:previous-work}. 
Visual CoT~\citep{Shao2024VisualCoTCA} relies on textual interpretations of images, limiting reasoning to the discrete text space. MCoT~\citep{Cheng2025VisualthoughtMCoTTA} enables continuous visual reasoning by editing or generating supplementary images, but requires substantial compute and lacks flexibility. VChain~\citep{huang2025vchain} interleaves images and text in the reasoning chain, yet still loses visual information by projecting images into text space.
\ours uniquely combines continuous visual reasoning, dense perceptual cues, and 3D-aware understanding within a single self-contained framework.

\begin{table*}[t]
\centering
\fontsize{7.5pt}{9pt}\selectfont

\newcommand{\colspace}{3pt}
\setlength{\tabcolsep}{\colspace}
\renewcommand\arraystretch{1.0}

\begin{tabular}{l|cccccccc}
\textbf{Desired Properties} 
& \textbf{VCoT} & \textbf{MCoT} & \textbf{VChain} & \textbf{Aurora} & \textbf{Mirage} & \textbf{LVR} & \textbf{Monet} & \textbf{Ours} \\
\Xhline{0.7pt}
\makecell[l]{Inferences without relying on external tools}
& \textcolor{ForestGreen}{\checkmark} 
& \textcolor{magenta}{\xmark} 
& \textcolor{ForestGreen}{\checkmark}
& \textcolor{ForestGreen}{\checkmark}
& \textcolor{ForestGreen}{\checkmark}
& \textcolor{ForestGreen}{\checkmark} 
& \textcolor{ForestGreen}{\checkmark} 
& \textcolor{ForestGreen}{\checkmark} \\

\hline
\makecell[l]{Reasons in the continuous visual space}
& \textcolor{magenta}{\xmark} 
& \textcolor{ForestGreen}{\checkmark} 
& \textcolor{ForestGreen}{\checkmark} 
& \textcolor{magenta}{\xmark} 
& \textcolor{ForestGreen}{\checkmark}
& \textcolor{ForestGreen}{\checkmark}
& \textcolor{ForestGreen}{\checkmark}
& \textcolor{ForestGreen}{\checkmark} \\

\hline
\makecell[l]{Leverages dense visual information for reasoning}
& \textcolor{magenta}{\xmark} 
& \textcolor{ForestGreen}{\checkmark} 
& \textcolor{magenta}{\xmark} 
& \textcolor{ForestGreen}{\checkmark} 
& \textcolor{magenta}{\xmark}
& \textcolor{magenta}{\xmark}
& \textcolor{magenta}{\xmark}
& \textcolor{ForestGreen}{\checkmark} \\

\hline
\makecell[l]{Has any type of perception enhancement}
& \textcolor{magenta}{\xmark} 
& \textcolor{magenta}{\xmark} 
& \textcolor{magenta}{\xmark} 
& \textcolor{ForestGreen}{\checkmark}
& \textcolor{magenta}{\xmark}
& \textcolor{magenta}{\xmark}
& \textcolor{magenta}{\xmark}
& \textcolor{ForestGreen}{\checkmark} \\

\end{tabular}
\vspace{+0.05cm}
\caption{
\textbf{Comparison of key properties with prior multimodal reasoning methods}. 
Unlike prior methods such as VCoT~\cite{Shao2024VisualCoTCA}, MCoT~\cite{zhangmultimodal}, VChain~\cite{huang2025vchain}, Aurora~\cite{bigverdi2025perception}, Mirage~\cite{li2025imaginereasoningspacemultimodal}, LVR~\cite{li2025latentLVR}, and Monet~\cite{wang2025monet}, \ours uniquely satisfies all desired properties: it reasons in continuous visual space, leverages dense visual cues, incorporates any type of perception, and operates fully without external tools.
\textcolor{ForestGreen}{Desired} and \textcolor{magenta}{undesired} properties are shown in \textcolor{ForestGreen}{green} and \textcolor{magenta}{magenta}, respectively.
}
\label{tab:previous-work}
\end{table*}

\noindent \textbf{Latent Space Reasoning}
Concurrent work shows that reasoning in latent space can strengthen LLMs in complex, multi-step tasks~\citep{Biran2024HoppingTL, Chen2024LanguageMA}. 
Coconut~\citep{Hao2024TrainingLL} finds that continuous latent embeddings are more efficient than explicit CoT, while CCoT~\citep{Cheng2024CompressedCO} compresses CoT into continuous tokens for denser reasoning. Other studies explore specialized reasoning tokens~\citep{Goyal2023ThinkBY} or use hidden states as implicit reasoning paths~\citep{Deng2023ImplicitCO}. 
Latent reasoning has also been extended to VLMs.
Aurora~\citep{bigverdi2025perception} employs VQ-VAE latents of depth and detection signals to enhance depth estimation and counting, whereas Mirage~\citep{yang2025machinementalimageryempower} uses latent imagination for visual reasoning tasks. LVR~\citep{li2025latentLVR} achieves reasoning in shared semantic and language space by reconstructing key visual tokens. Monet~\citep{wang2025monet} introduces a novel reinforcement learning algorithm to optimize latent visual reasoning.
Different from these works (shown in Tab.~\ref{tab:previous-work}), \ours introduces a form of implicit tool use directly embedded in dense continuous latent space, where the implicit `tools' are arbitrary types of visual thinking tokens tied to their specific perceptual experts.

\section{\oursfull (\ours)}
\label{sec:method}
We first introduce the preamble in \cref{sec:method-preamble}. 
We then show the overall pipeline of \ours in \cref{sec:method-framework}. 
We also discuss how we select the visual token categories and explain how different visual tokens are aligned (\cref{sec:method-align}). 
Finally, we present the model training pipeline, \eg, the training loss formulation and the data framework design in (\cref{sec:method-loss}).

\subsection{Preamble}
\label{sec:method-preamble}
Existing VLMs face two key limitations in fine-grained visual reasoning. 
\textbf{1) \textit{Text-only CoT accumulates errors.}} Text-only CoT executes a long chain of thought, which may generate errors at the early stage. These mistakes will accumulate and ultimately lead to an incorrect final result. Therefore, we need a reasoning that is short and effective.
\textbf{2) \textit{Supervision is dominated by text responses,}} which provides little incentive for the model to capture \textit{low-level perceptual cues} such as edges, depth, or regions.
We need to equip VLMs themselves with the capability of extracting fine-grained visual information from the image, which can be further decoded by vision decoders.

\ours intends to provide a foundation for the next generation of multimodal reasoning systems, capable of thinking fluidly across both language and vision in a self-contained, interpretable manner.

\subsection{\ours Overall Pipeline}
\label{sec:method-framework}

We propose \ours, a framework that augments VLMs with \textbf{\textit{chains of visual thoughts}}. Fig.~\ref{fig:method-framework} illustrates the overview of \ours pipeline. 
Essentially, this framework equips VLMs with the capability of outputting fine-grained visual representations within a continuous visual token space, enabling them to reason directly over rich perceptual information and maintain spatial and geometric coherence throughout the reasoning process.

\begin{figure*}[t]
    \centering
    \includegraphics[width=1.0\linewidth]{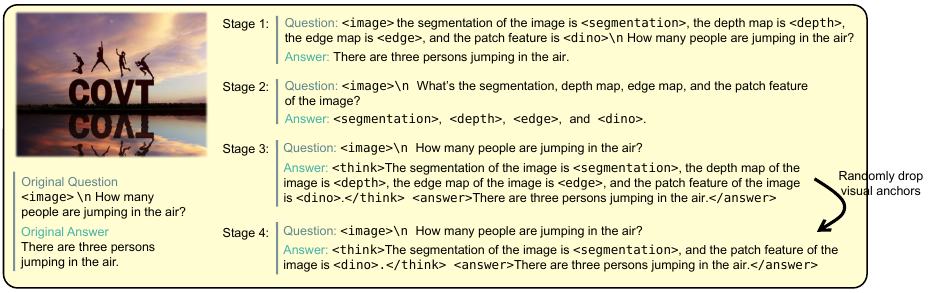}
    \caption{\textbf{Four-stage data formatting for \ours.} The first stage helps the model comprehend the visual tokens, and the second stage guides it to generate them. The third stage enables the VLM to integrate visual tokens into its reasoning process, while the final stage allows the model to efficiently select and utilize visual thinking tokens within visual thought chains.}
    \vspace{-0.2cm}
    \label{fig:data}
\end{figure*}

At its core, \ours retains the standard next-token prediction paradigm.
For standard VLMs, given visual features $\mathcal{V}$ extracted from a frozen vision encoder and text features $\mathcal{T}$ from a language encoder, the VLM estimates the probability of generating a sequence $Y\!=\!(y_1, y_2, \ldots, y_n)$ as:
\begin{align}
P(Y \mid \mathcal{V}, \mathcal{T}; \theta) = \prod_{i=1}^n P\left(y_i \mid y_{<i}, \mathcal{V}, \mathcal{T}\right).
\vspace{-3pt}
\end{align}
As shown in Fig.~\ref{fig:method-framework}, \ours extends this formulation by introducing \emph{Chain-of-Visual-Thought tokens}, where each token $y_i$ can represent either a visual token or a text token. 


To effectively incorporate \ours tokens into the VLM, we train the model to function as a dense visual encoder capable of generating multiple visual tokens that capture diverse fine-grained perceptual cues. 
The VLM is trained to generate CoVT tokens that, through task-specific decoders, reconstruct visual outputs under reconstruction supervision. Through this process, \ours evolves to generate rich, fine-grained visual information across multiple perceptual dimensions within a thinking chain.

\subsection{\ours Tokens}
\label{sec:method-align}

\textbf{Token selection based on core perception ability.} As proposed in ~\citep{Zhou2025FromPTsurvey}, the vision-centric perceptual ability of VLMs can be summarized as \textit{(i) instance recognition}, \textit{(ii) 2D and 3D spatial relationships}, \textit{(iii) structure detection}, and \textit{(iv) deep mining of semantic information}. 
Based on this categorization, we use four vision models to supervise \ours tokens to learn each ability:
\textit{1) Segmentation tokens provide instance-level position and shape information}, which endow VLMs with the instance recognition signals and 2D spatial perception. 
\textit{2) Depth tokens provide pixel-level depth information}, equipping VLMs with the capability of figuring out 3D spatial relationships. 
\textit{3) Edge tokens provide geometry-level details}, which assist models to detect structural cues and partially provide 2D spatial information.
\textit{4) DINO tokens provide the patch-level representation of the images}, delivering rich semantic information.


\noindent \textbf{Tokens alignment based on granularity of visual models.} 
Task-oriented models and representative models produce outputs at different levels of granularity. In general, task-oriented models tend to be more fine-grained, whereas representative models are usually less fine-grained.
We adopt different strategies to align each type of token with the visual models based on different granularities.
Essentially, we adopt two main alignment methods: For fine-grained task-oriented models, visual tokens are projected to the prompt space and then aligned at the prompt level with the decoders, while for representative models, alignment with the encoders is applied at the feature level after the projection. The projection layer consists of one fully connected layer and one multi-head attention layer.

\textit{1) Segmentation tokens are supervised by SAM}~\citep{Kirillov2023SegmentASAM}, which is a \textit{task-oriented model} that contains dense visual features. 
Therefore, following LISA~\citep{Lai2023LISARS}, we align Segmentation tokens with the SAM decoder. 
The 8 Segmentation tokens are aligned at the prompt level, and each token prompts one mask, formulated as:
\begin{equation}
    \hat{M}_i = \text{Decoder}(T_i^{\text{sam}}, f), \quad \hat{M}_i \in [0,1]^{H \times W},
\end{equation}
where $\hat{M}_i$ is the $i$th decoded mask, $T_i^{\text{sam}}$ denotes the $i$th segmentation token predicted in \ours, serving as the prompt fed into the SAM decoder, and $f$ means the dense embedding from the SAM encoder.
During the training process, the Hungarian matching algorithm is employed to match the predicted masks with the ground truths, while dice loss and focal loss are applied.

\textit{2) Depth tokens are supervised by DepthAnything v2}~\citep{Yang2024DepthAVDepthAnything}\textit{--a task-oriented model}. Since these tokens contain dense information, they are also aligned with the decoder at the prompt level. 
We use 4 Depth tokens to serve as 4 prompts to interact with the dense features extracted by DepthAnythingv2 through batch matrix multiplication (BMM) to reconstruct the depth map, formulated as:
\begin{equation}
    \hat{D}_i = \text{softmax}\left( T_{i}^{\text{depth}} \cdot F_i^{\text{depth}\top} \right),
\end{equation}
where $\hat{D}_i$ denotes the $i$th reconstructed depth map, $T_{i}^{\text{depth}}$ represents the $i$th depth visual token, and $F_i^{\text{depth}}$ is the $i$th middle layer feature from DepthAnything encoder. The final depth map is $\hat{D} = \frac{\sum_{i=0}^{3} \hat{D}_i}{4}.$
The L1 reconstruction loss is employed for aligning Depth tokens.

\textit{3) Edge tokens are aligned with PIDINet}~\citep{Su2021PixelDNPIDINet}. Each of 4 Edge tokens functions as an $1 \times 1$ convolutional kernel applied to the dense features from PIDINet encoder to reconstruct the $i$th edge map $\hat{E}_i$. The final edge map is $\hat{E} = \frac{\sum_{i=0}^3 \hat{E}_i}{4}$, and aligned via L1 loss function.

\textit{4) DINO tokens are supervised by DINOv2}~\citep{Oquab2023DINOv2LR}, which is trained as the \textit{representative model}, extracting patch-level features. Therefore, the 4 DINO tokens are mapped into the same shape as the DINO feature using the projection layer, and aligned under an MSE objective.

\begin{table*}[th]
\centering
\fontsize{9pt}{9.2}\selectfont
\setlength{\tabcolsep}{3.5pt}
\renewcommand\arraystretch{0.98}
\scriptsize
\newcommand{\pos}[1]{\textcolor{green!60!black}{#1}}
\newcommand{\negval}[1]{\textcolor{red!70!black}{#1}}
\begin{tabular}{ccccccccccccccccc}
\toprule
\multicolumn{4}{c}{\textbf{Visual tokens}} & 
\multicolumn{4}{c}{\textbf{CV-Bench}} &
\multicolumn{9}{c}{\textbf{Other vision-centric benchmarks}} \\
\cmidrule(lr){1-4}
\cmidrule(lr){5-8}
\cmidrule(lr){9-17}
Seg & Depth & DINO & Edge & CVBench & Count & Depth & Dist. & BLINK & RW-QA & MMT & MMStar-P & MMVP & MME-RW & V* & $\text{HR}_{4K}$ & $\text{HR}_{8K}$ \\
\midrule
\multicolumn{17}{l}{\textcolor{gray}{\textbf{Closed-source Models}}} \\
\multicolumn{4}{l}{\textcolor{gray}{Claude-4-Sonnet}} & \textcolor{gray}{76.3} & \textcolor{gray}{62.2} & \textcolor{gray}{77.7} & \textcolor{gray}{80.5} & \textcolor{gray}{39.6} & \textcolor{gray}{63.7} & \textcolor{gray}{-} & \textcolor{gray}{58.8} & \textcolor{gray}{48.7} & \textcolor{gray}{-} & \textcolor{gray}{15.2} & \textcolor{gray}{32.3} & \textcolor{gray}{22.7} \\
\multicolumn{4}{l}{\textcolor{gray}{GPT-4o}} & \textcolor{gray}{79.2} & \textcolor{gray}{65.6} & \textcolor{gray}{86.7} & \textcolor{gray}{81.0} & \textcolor{gray}{63.0} & \textcolor{gray}{69.7} & \textcolor{gray}{-} & \textcolor{gray}{65.2} & \textcolor{gray}{72.0} & \textcolor{gray}{-} & \textcolor{gray}{42.9} & \textcolor{gray}{50.6} & \textcolor{gray}{46.7} \\
\midrule
\multicolumn{4}{l}{Qwen2.5-VL-7B} & 74.5 & 65.0 & 72.8 & 75.5 & 55.7 & 68.6 & 61.7 & 67.1 & 56.0 & 60.0 & 76.4 & 68.6 & 64.9 \\
\multicolumn{4}{l}{LVR~\citep{li2025latentLVR}$^{\dag}$} & 77.2 & 67.0 & 86.2 & 71.2 & 55.9 & 69.7 & -- & 67.7 & -- & -- & 81.7 & 69.6 & 65.6 \\
\multicolumn{4}{l}{Monet~\citep{wang2025monet}} & 75.7$^{\dag}$ & \textbf{68.8}$^{\dag}$ & 77.7$^{\dag}$ & 73.8$^{\dag}$ & 52.9$^{\dag}$ & 69.3$^{\dag}$ & -- & 67.1$^{\dag}$ & -- & -- & \textbf{83.3} & 71.0 & 68.0 \\
\midrule
\multicolumn{17}{l}{\textbf{\ours (1 Visual Token)}} \\
\rowcolor{yellow!10}
\checkmark &   &   &   & 77.9 & 66.0 & 80.8 & 80.5 & \textbf{57.4} & 71.1 & 62.1 & 68.5 & 58.7 & 62.1 & 79.1 & 71.9 & 69.0 \\
\rowcolor{yellow!10}
 & \checkmark &   &   & 78.7 & 65.4 & 83.2 & 78.2 & 56.4 & 71.5 & \textbf{62.7} & \textbf{69.9} & 58.7 & 62.0 & 79.1 & 71.9 & 69.4 \\
\rowcolor{yellow!10}
 &   & \checkmark &   & 71.3 & 64.7 & 72.3 & 66.7 & 55.8 & 71.5 & 62.5 & 67.9 & 57.3 & 61.1 & 77.5 & 71.0 & 68.6 \\
\midrule
\multicolumn{17}{l}{\textbf{\ours (3 Visual Tokens)}} \\
\rowcolor{yellow!10}
\checkmark & \checkmark & \checkmark &   & \textbf{80.0} & 66.2 & 86.8 & \textbf{82.5} & 56.0 & 71.6 & 62.1 & 69.2 & \textbf{58.7} & \textbf{63.7} & 78.0 & \textbf{72.9} & 69.4 \\
\rowcolor{yellow!10}
\multicolumn{4}{l}{\textit{$\Delta$ (\textit{vs} Baseline)}} & \textbf{\pos{+5.5}} & \textbf{\pos{+1.2}} & \textbf{\pos{+14.0}} & \textbf{\pos{+7.0}} & \textbf{\pos{+0.3}} & \textbf{\pos{+3.0}} & \textbf{\pos{+0.4}} & \textbf{\pos{+2.1}} & \textbf{\pos{+2.7}} & \textbf{\pos{+3.7}} & \textbf{\pos{+1.6}} & \textbf{\pos{+4.3}} & \textbf{\pos{+4.5}} \\
\midrule
\multicolumn{17}{l}{\textbf{\ours (4 Visual Tokens)}} \\
\rowcolor{yellow!10}
\checkmark & \checkmark & \checkmark & \checkmark & 79.8 & 66.1 & \textbf{89.2} & 80.5 & 56.2 & \textbf{71.8} & 61.9 & 68.4 & 56.7 & 63.3 & 78.5 & 72.5 & \textbf{69.9}  \\
\rowcolor{yellow!10}
\multicolumn{4}{l}{$\Delta$ (\textit{vs Baseline})} & \textbf{\pos{+5.3}} & \textbf{\pos{+1.1}} & \textbf{\pos{+16.4}} & \textbf{\pos{+5.0}} & \textbf{\pos{+0.5}} & \textbf{\pos{+3.2}} & \textbf{\pos{+0.2}} & \textbf{\pos{+1.3}} & \textbf{\pos{+0.7}} & \textbf{\pos{+3.3}} & \textbf{\pos{+2.1}} & \textbf{\pos{+3.9}} & \textbf{\pos{+5.0}} \\
\bottomrule
\end{tabular}
\vspace{-0.1cm}
\caption{\textbf{Comparison of \ours with the baseline, closed-source models, and recent latent visual reasoning methods.} \ours delivers consistent improvements across vision-centric benchmarks and further reveals that each visual token type contributes most effectively to tasks related to its rich information. $^{\dag}$ indicates our reproduced results based on the released checkpoints.
}
\label{tab:final_exp1}
\end{table*}

\subsection{\ours Training}
\label{sec:method-loss}
\noindent \textbf{Training Loss.}
During training, 
the \textit{joint loss function} is defined as:
\begin{align}
\mathcal{L}_{\text{total}} 
&= \mathcal{L}_{\text{ce}} 
  + \gamma \big(
      \lambda_{\text{seg}} \cdot \mathcal{L}_{\text{visual}}^{\text{seg}} 
    + \lambda_{\text{depth}} \cdot \mathcal{L}_{\text{visual}}^{\text{depth}} \nonumber\\
  &\phantom{= \mathcal{L}_{\text{ce}} + \gamma (}
    + \lambda_{\text{edge}} \cdot \mathcal{L}_{\text{visual}}^{\text{edge}} 
    + \lambda_{\text{dino}} \cdot \mathcal{L}_{\text{visual}}^{\text{dino}}
  \big),
\label{eq:total-loss}
\end{align}
where $\mathcal{L}_{\mathrm{ce}}$ is the typical cross-entropy loss for VLMs, $\gamma$ is the coefficient of visual loss, and all of the $\lambda$ coefficients are the weighting factors for the losses of the corresponding visual tasks.
During the inference process, the visual thinking tokens are not decoded.

Additionally, our framework supports the flexible integration of new visual token types. 
Because the pipeline follows a clean next-token prediction paradigm, additional tokens can be incorporated with minimal modification.


\newcommand{\BaseOverall}{55.7}
\newcommand{\BaseCounting}{70.8}
\newcommand{\BaseLoc}{54.1}
\newcommand{\BaseDepth}{81.5}
\newcommand{\BaseCorr}{52.9}
\newcommand{\BaseSim}{86.7}

\newcommand{\BaseCVBenchOverall}{75.1}
\newcommand{\BaseCVBenchCount}{65.2}
\newcommand{\BaseCVBenchDepth}{72.8}
\newcommand{\BaseCVBenchDistance}{72.8}

\newcommand{\BaseRealWorldQA}{68.6}
\newcommand{\BaseMMT}{61.7}
\newcommand{\BaseMMStar}{67.1}
\newcommand{\BaseMME}{60.0}
\newcommand{\BaseMMVP}{56.0}
\newcommand{\BaseVS}{76.4}
\newcommand{\BaseOCR}{88.5}
\newcommand{\BaseFourK}{68.6}
\newcommand{\BaseEightK}{64.9}


\noindent \textbf{Training Data.}
To enable VLMs to progressively learn the visual tokens while not losing ability in the text space, \ours introduces four data formatting stages as shown in Fig.~\ref{fig:data}.
This guides the VLMs to learn progressively through the sequence from understanding visual tokens \textit{(comprehension stage)}, to generating visual tokens \textit{(generation stage)}, to reasoning with chain of visual thoughts \textit{(reasoning stage)}, and finally to dynamically using visual tokens in the thinking chain \textit{(efficient reasoning stage)}.

In \textit{1) comprehension stage}, we insert visual tokens after \texttt{<image>} to teach the VLMs to learn the basic semantics of the visual tokens. 
In \textit{2) generation stage}, we modify the question and answer, as shown in Fig. ~\ref{fig:data}, to guide the VLMs to generate the visual tokens precisely.
\textit{3) reasoning stage} introduces the chain-of-visual-thought format, where the visual tokens are used within the thinking process. This teaches the model to leverage the visual tokens to derive the final answers.
\textit{4) efficient reasoning stage} randomly drops out some sets (ranging from $0$ to $k$, where $k$ is the number of token types) of visual tokens. With a portion of visual token types, \ours learns to utilize all features effectively rather than being constrained by a fixed output pattern.

The dataset used for training includes: (1) a vision-centric (and also real-world) subset of the LLaVA-OneVision dataset~\citep{Li2024LLaVAOneVisionEV}.
(2) spatial perception data, including TallyQA~\citep{Acharya2018TallyQAAC} and ADE20K-Depth~\citep{Zhou2016SemanticUOADE, bigverdi2025perception}.

\begin{figure*}[t]
    \centering
    \includegraphics[width=1.0\linewidth]{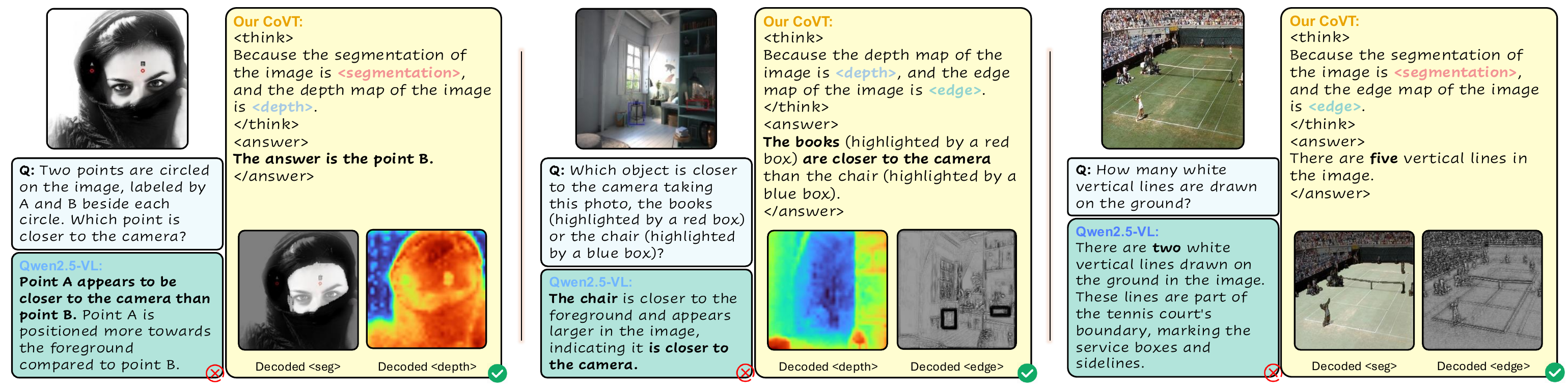}
    \vspace{-0.65cm}
    \caption{\textbf{Visualization of \ours tokens.} Different visual tokens contribute complementary cues that enable the model to solve complex perceptual reasoning tasks. \textbf{Left:} Segmentation tokens localize point B on the face, while the depth tokens capture the relative depth relationships. \textbf{Mid:} Depth visual tokens provide depth map information, and the edge tokens help highlight the positions of two boxes. \textbf{Right:} The Segmentation tokens identify the attended region, and the edge tokens delineate the fine-grained line structures.}
    \label{fig:vis-demo2}
    \vspace{-0.1cm}
\end{figure*}


\section{Experiments}


In the experiment section, we first describe the experimental settings of \oursfull (\ours) in \cref{sec:exp-detail}. 
Second, we introduce the benchmarks results on both vision-centric and non-vision-centric datasets in \cref{sec:exp-eval}. 
Third, we present the quantitative results demonstrating the advantages of \ours in \cref{sec:exp-result}. 
Finally, we ``visualize'' the continuous visual tokens in \cref{sec:exp-vis} and conduct ablation studies in \cref{sec:exp-ablat}.


\subsection{Experiment Details}
\label{sec:exp-detail}

In our experiments, Qwen2.5-VL-7B~\citep{Qwen2.5-VL} is selected as the main baseline, \ours uses LoRA~\cite{Hu2021LoRALA} tuning method, while the rank of LoRA is 16 and the LoRA alpha is 32. The learning rate of LoRA is set as $5\times 10^{-5}$ and the projection layer learning rate is set to $1\times 10^{-5}$. Batch size is set to 4. For one type of visual tokens, the first phase steps are 4000, second and third phase steps are 3000, and the fourth phase steps are 4000. For three types of visual tokens, the first phase steps are 6000, second and third phase steps are 3000, and the fourth phase steps are 5000. For four types of visual tokens, first phase steps are 8000, second and third phase steps are 3000, and the fourth phase steps are 6000. The experiments are carried out on 1$\times$A100 or 4$\times$A6000 GPUs. $\gamma$ and all of the $\lambda$ in Eq.~\ref{eq:total-loss} are set as 1.


\begin{table}[t!]
\centering
\fontsize{7.7pt}{9.2}\selectfont

\newcommand{\colspace}{2.5pt}
\newcommand{\pos}[1]{\textcolor{green!60!black}{#1}}
\newcommand{\negval}[1]{\textcolor{red!70!black}{#1}}
\newcommand{\eqlval}[1]{\textcolor{gray!100!black}{#1}}
\setlength{\tabcolsep}{\colspace}
\renewcommand\arraystretch{0.95}

\begin{tabular}{lcccccccc}
\toprule
&
\multicolumn{3}{c}{\textbf{CV-Bench}}
& \multicolumn{5}{c}{\textbf{BLINK}} \\
\cmidrule(lr){2-4}\cmidrule(lr){5-9}
& Count & Depth & Dist.
& Count & \makecell[c]{Obj.\\Loc.}
& \makecell[c]{Rel.\\Depth}
& \makecell[c]{Vis.\\Corr.}
& \makecell[c]{Vis.\\Sim.} \\
\midrule

LLaVA
& 59.3 & 61.8 & 50.2
& 56.7 & 54.9 & 52.4
& 29.7 & 51.1 \\
\midrule

Aurora$^{\dag}$ (\textit{depth})
& 54.9 & 67.7 & \textbf{52.3}
& 53.3 & 55.7 & 62.9
& 26.2 & 47.4 \\

\rowcolor{yellow!10}
\ours (\textit{w/ Depth})
& 60.7 & \textbf{71.0} & \textbf{52.3}
& 56.7 & \textbf{59.8} & \textbf{75.8}
& \textbf{31.4} & \textbf{53.3} \\

\rowcolor{yellow!10}
$\Delta$ (\textit{vs Aurora}) & \textbf{\pos{+5.8}} & \textbf{\pos{+3.3}} & \textbf{\eqlval{+0.0}} & \textbf{\pos{+3.4}} & \textbf{\pos{+4.1}} & \textbf{\pos{+12.9}} & \textbf{\pos{+5.2}} & \textbf{\pos{+5.9}} \\
\midrule

Aurora$^{\dag}$ (\textit{count})
& 56.0 & 62.2 & 47.8
& 31.7 & 26.2 & 24.2
& 26.7 & 21.5 \\

\rowcolor{yellow!10}
\ours (\textit{w/ Seg})
& \textbf{61.9} & 60.7 & 51.3
& \textbf{58.3} & 56.6 & 69.4
& 29.7 & 52.6 \\

\rowcolor{yellow!10}
$\Delta$ (\textit{vs Aurora}) & \textbf{\pos{+5.9}} & \textbf{\negval{-1.5}} & \textbf{\pos{+3.5}} & \textbf{\pos{+26.6}} & \textbf{\pos{+30.4}} & \textbf{\pos{+45.2}} & \textbf{\pos{+3.0}} & \textbf{\pos{+31.1}} \\
\bottomrule

\end{tabular}
\vspace{-0.1cm}
\caption{\textbf{Comparison between \ours and Aurora based on LLaVA-v1.5-13B.}
$^{\dag}$ indicates our reproduced results based on the provided checkpoints.}
\vspace{-0.1cm}
\label{tab:final_exp_2}
\end{table}

\subsection{Model Evaluation}
\label{sec:exp-eval}

All evaluations are performed using VLMEvalKit~\citep{Duan2024VLMEvalKitAO}.

\noindent \textbf{Vision-centric benchmarks.} Our main focus is on CV-Bench. In particular, from CV-Bench we highlight the sub-tasks \emph{Count}, \emph{Depth}, and \emph{Distance}. These sub-tasks act as precise indicators to validate the effectiveness of our method. 
We further evaluate on other vision-centric benchmarks, including BLINK~\citep{Fu2024BLINKML}, RealWorldQA (RW-QA)~\citep{Grok15vpreview}, MMT-Bench (MMT)~\citep{Ying2024MMTBenchAC}, MMStar~\citep{Chen2024AreWOMMStar}, MMVP~\citep{Tong2024EyesWSMMVP}, MME-RealWorld (MME-RW)~\citep{Zhang2024MMERealWorldCY}, V* Bench (V*)~\citep{Wu2023VGVStar}, and HRBench (HR$_{4K}$ and HR$_{8K}$)~\citep{Wang2024DivideCAHRBench}. Among them, for MMStar we specifically choose the \emph{Coarse Perception}, \emph{Fine-grained Perception}, and \emph{Instance Reasoning} subsets (MMStar-P), as they are more aligned with real-world reasoning.

\noindent \textbf{Non-vision-centric benchmarks.} Besides, we also evaluate \ours on some non-vision-centric visual benchmarks such as OCRBench~\citep{Liu2023OCRBenchOT},  MME~\citep{Fu2023MMEAC}, MUIRBench~\citep{Wang2024MuirBenchAC}, HallusionBench~\citep{Guan2023HallusionbenchAA}, A-OKVQA~\citep{Schwenk2022AOKVQAAB}, TaskMeAnything~\citep{Zhang2024TaskMA}, WeMATH~\citep{Qiao2024WeMathDY}, and WorldMedQA-V~\citep{Matos2024WorldMedQAVAM}. For MME, we select the text-centric sub-task \textit{text translation} as the evaluation of the text-centric performance. 

\subsection{Quantitative Results}
\label{sec:exp-result}


\noindent \textbf{\ours outperforms the baseline across the vision-centric benchmarks.} As shown in Tab.~\ref{tab:final_exp1}, \ours is capable of incorporating various kinds of visual tokens. We employ three visual tokens, Segmentation, Depth, and DINO, as our main results. Compared to the baseline, \ours consistently achieve large gains across the main vision-centric benchmarks. \ours improves by 5.5\% on CV-Bench, 14.0\% on the subtask \textit{depth} in CV-Bench, 3.7\% on MME-RealWorld, and 4.5\% on HRBench8K. These results indicate \ours with visual thinking chain improves across visual-centric and fine-grained perceptual tasks.

\noindent \textbf{\ours generalizes to the other baseline.} In addition to the experiment based on Qwen, \ours is also implemented based on LLaVA-v1.5-13B, in order to compare \ours with Aurora. As shown in Fig.~\ref{tab:final_exp_2}, for \ours with depth tokens, it excels Aurora-\textit{depth} by 12.9\% on \textit{relative-depth} in BLINK. For \ours using segmentation \ours tokens, outperforms Aurora-\textit{count} by 26.6\% on BLINK-\textit{count} benchmark. These results indicate that \ours generalizes on various baselines across vision-centric tasks.

\subsection{Qualitative Results}
\label{sec:exp-vis}


To better understand why \ours is effective, we select several examples and decode the \ours tokens from the model outputs to visualize whether these tokens provide useful information for reasoning.



Fig.~\ref{fig:vis-demo2} illustrates that different \ours tokens carry different rich fine-grained information, and the cues they provide are highly complementary. To be interpretable, \ours tokens are decoded into fine-grained output (\textit{e.g.} masks, depth maps, edge maps). \textbf{\textit{For the left example}}, the Segmentation token provides 2D perceptual cues by localizing ``point B'' on the face, while the Depth token supplies 3D information, indicating that the face region is closer to the camera than the surrounding areas.
\textbf{\textit{For the middle example}}, rhe Depth token encodes depth perception, whereas the Edge token supplies fine-grained boundary cues for the two target objects.
This example is from \textit{Depth} sub-task in CV-Bench, thus the synergy explains the 2.4\% improvement observed when \ours uses four visual tokens instead of three on the CVBench-Depth task, as shown in Tab.~\ref{tab:final_exp1}.
\textbf{\textit{For the right example}}, the Segmentation token localizes the target region and the Edge token emphasize fine-grained boundaries, which is difficult for the Segmentation token, deriving the correct answer through chains of visual thoughts.

\begin{figure}
    \centering
    \includegraphics[width=1.05\linewidth]{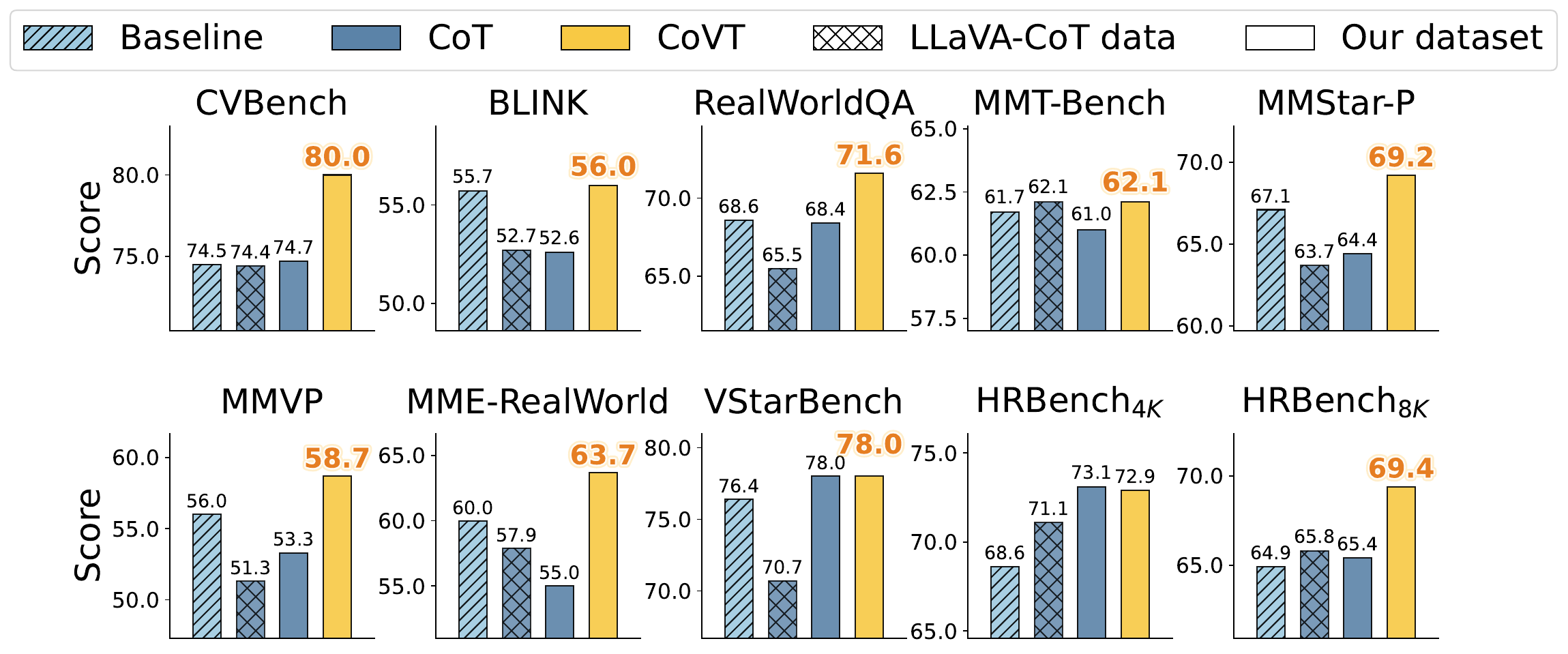}
    \caption{\textbf{Text-only CoT \textit{vs} \ours.} \ours substantially enhances VLMs' capabilities on vision-centric tasks, whereas text-only CoT can even degrade performance.}
    \label{fig:exp-cot-vs-covt}
    \vspace{-0.4cm}
\end{figure}

\begin{table}[t]
\centering
\fontsize{8.3pt}{10pt}\selectfont
\setlength{\tabcolsep}{2.0pt}
\renewcommand\arraystretch{0.95}
\newcommand{\pos}[1]{\textcolor{green!60!black}{#1}}
\newcommand{\negval}[1]{\textcolor{red!70!black}{#1}}
\begin{tabular}{lcccccccc}
\toprule
Quantity & CVBench & BLINK & RW-QA & MM*-P & MMVP & V* & $\text{HR}_{4K}$ \\
\midrule
0 token     & 76.6 & 55.5 & 70.7 & 68.0 & 55.3 & 77.5 & 68.6 \\
16 empty & 75.7 & \textbf{56.0} & 70.3 & 67.9 & 56.7 & 77.5 & 68.1 \\
\midrule
1 token     & 78.9 & 55.6 & 70.8 & 68.8 & 56.7 & \textbf{78.5} & \textbf{73.0} \\
\rowcolor{yellow!12}
8 tokens     & \textbf{80.0} & \textbf{56.0} & \textbf{71.6} & \textbf{69.2} & \textbf{58.7} & 78.0 & 72.9 \\
32 tokens    & 73.9 & 54.4 & 68.4 & 62.1 & 55.3 & 77.2 & 70.8 \\
\bottomrule
\end{tabular}
\vspace{-0.1cm}
\caption{\textbf{\ours Ablation on segmentation token numbers.} The appropriate token number is essential for \ours. \textbf{0 token} is the control group (direct fine-tuning), while \textbf{16 empty tokens} serve to isolate the role of the token embeddings themselves. 8 segmentation tokens perform the best.}
\vspace{-0.1cm}
\label{tab:ablation1}
\end{table}

\subsection{Ablation Studies}
\label{sec:exp-ablat}


\noindent \textbf{Text-only Chain-of-Thought \vs Chain-of-Visual-Thought.}
To isolate the contribution of continuous visual tokens, we conduct an ablation comparing text-only CoT with our full \ours framework. 
For the text-only setting, we follow the CoT formatting paradigm used in the LLaVA-CoT 100k dataset and apply the same formatting to our own training data, ensuring full consistency with \ours except for the absence of visual tokens. 
Fig.~\ref{fig:exp-cot-vs-covt} shows that text-only CoT not only fails to improve performance on vision-centric reasoning tasks, but often degrades it. 
In contrast, \ours consistently enhances performance across vision-centric benchmarks, highlighting the necessity of continuous visual tokens for effective visual reasoning.


\begin{table}[t]
\centering
\fontsize{7.8pt}{9.5pt}\selectfont
\setlength{\tabcolsep}{2.0pt}
\renewcommand\arraystretch{0.95}

\begin{tabular}{l lccccccccc}
\toprule
Type & Align & CVBench & BLINK & RW-QA & MM*-P & MMVP & V* & $\text{HR}_{4K}$ \\
\midrule
\multirow{2}{*}{Seg} 
& Feature    & 76.8 & 55.2 & 70.6 & 67.7 & 56.0 & 78.0 & 69.8 \\
& \cellcolor{yellow!12}Ours   
  & \cellcolor{yellow!12}\textbf{77.9} 
  & \cellcolor{yellow!12}\textbf{57.4} 
  & \cellcolor{yellow!12}\textbf{71.1} 
  & \cellcolor{yellow!12}\textbf{68.5} 
  & \cellcolor{yellow!12}\textbf{58.7} 
  & \cellcolor{yellow!12}\textbf{79.1} 
  & \cellcolor{yellow!12}\textbf{71.9} \\

\midrule

\multirow{2}{*}{Depth} 
& Feature    & 77.0 & 54.2 & 70.5 & 67.6 & 55.3 & \textbf{78.0} & 71.3 \\
& \cellcolor{yellow!12}Ours   
  & \cellcolor{yellow!12}\textbf{78.7} 
  & \cellcolor{yellow!12}\textbf{56.4} 
  & \cellcolor{yellow!12}\textbf{71.5} 
  & \cellcolor{yellow!12}\textbf{69.9} 
  & \cellcolor{yellow!12}\textbf{58.7} 
  & \cellcolor{yellow!12}77.5 
  & \cellcolor{yellow!12}\textbf{71.9} \\

\bottomrule
\end{tabular}

\vspace{-0.1cm}
\caption{Our tailored \textbf{alignment strategy} plays a crucial role in further enhancing the performance of \ours.}
\vspace{-0.2cm}
\label{tab:sam_depth_compare}
\end{table}

\noindent \textbf{Token Numbers.} 
We ablate various numbers of segmentation visual tokens, as shown in Tab.~\ref{tab:ablation1}. 
The ``0 token'' setting corresponds to directly fine-tuning the base model on our dataset. 
The ``16 empty'' setting replaces our 16 visual thinking tokens with 16 ordinary tokens without any visual alignment, serving as a pure latent-reasoning baseline.
Settings with 1, 8, and 32 Segmentation tokens vary the token count while keeping the Depth and DINO tokens fixed at 4 each; the 8-token setting corresponds to our full model. 

We observe that using too few Segmentation tokens leads to performance degradation, though still better than the ``0 token'' baseline. 
However, increasing the token count to 32 harms performance, maybe due to the difficulty of aligning a large number of segmentation tokens.
The poor performance of the ``16 empty'' variant further highlights the importance of visually aligned tokens.
Overall, the results demonstrate that 8 Segmentation tokens, together with 4 Depth and 4 DINO tokens, form a balanced and effective configuration, and that visual alignment is essential for enhancing vision-centric perception in VLMs.

\begin{figure}[H]
    \centering
    \includegraphics[width=0.99\linewidth]{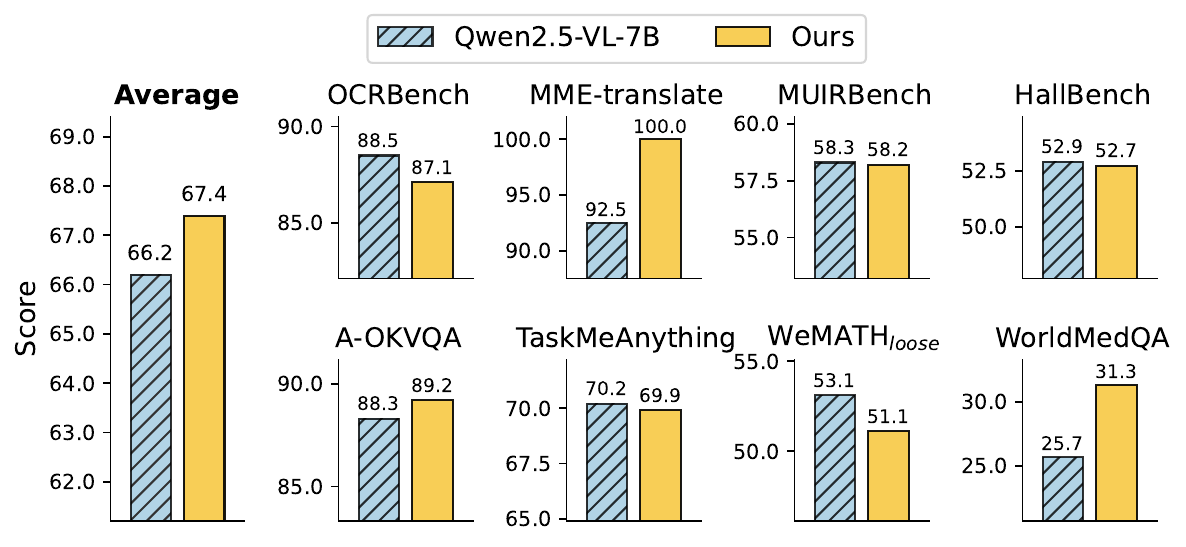}
    \vspace{-0.2cm}
    \caption{Beyond the gains on vision-centric benchmarks, \ours also achieves slight improvements on \textbf{non–vision-centric tasks}}
    \vspace{-0.4cm}
    \label{fig:final_exp3}
\end{figure}

\noindent \textbf{Segmentation and Depth Alignment Strategies.} 
We ablate two alignment approaches. 
Our primary method aligns \ours{} tokens through task decoders, enabling the tokens to capture richer and more fine-grained perceptual cues. 
In contrast, direct feature alignment applies an MSE loss between the visual tokens and the encoder features of the corresponding visual model, which inevitably loses important perceptual details from the image.

As shown in Tab.~\ref{tab:sam_depth_compare}, direct feature alignment consistently underperforms \ours. 
These results highlight the importance of our tailored alignment strategies and demonstrate that aligning visual tokens with decoders yields more effective and perceptually grounded representations.

\noindent \textbf{\ours remains competitive across various non-vision-centric benchmarks.} Fig.~\ref{fig:final_exp3} shows our method remains comparable performance, with 1.2\% improvement over eight non-vision-centric benchmarks, demonstrating that \ours does not lead to a notable degradation in the generalization, and even yields an improvement for overall.

\begin{strip}
\centering

\begin{minipage}{0.5\textwidth}
\centering
\includegraphics[width=\linewidth]{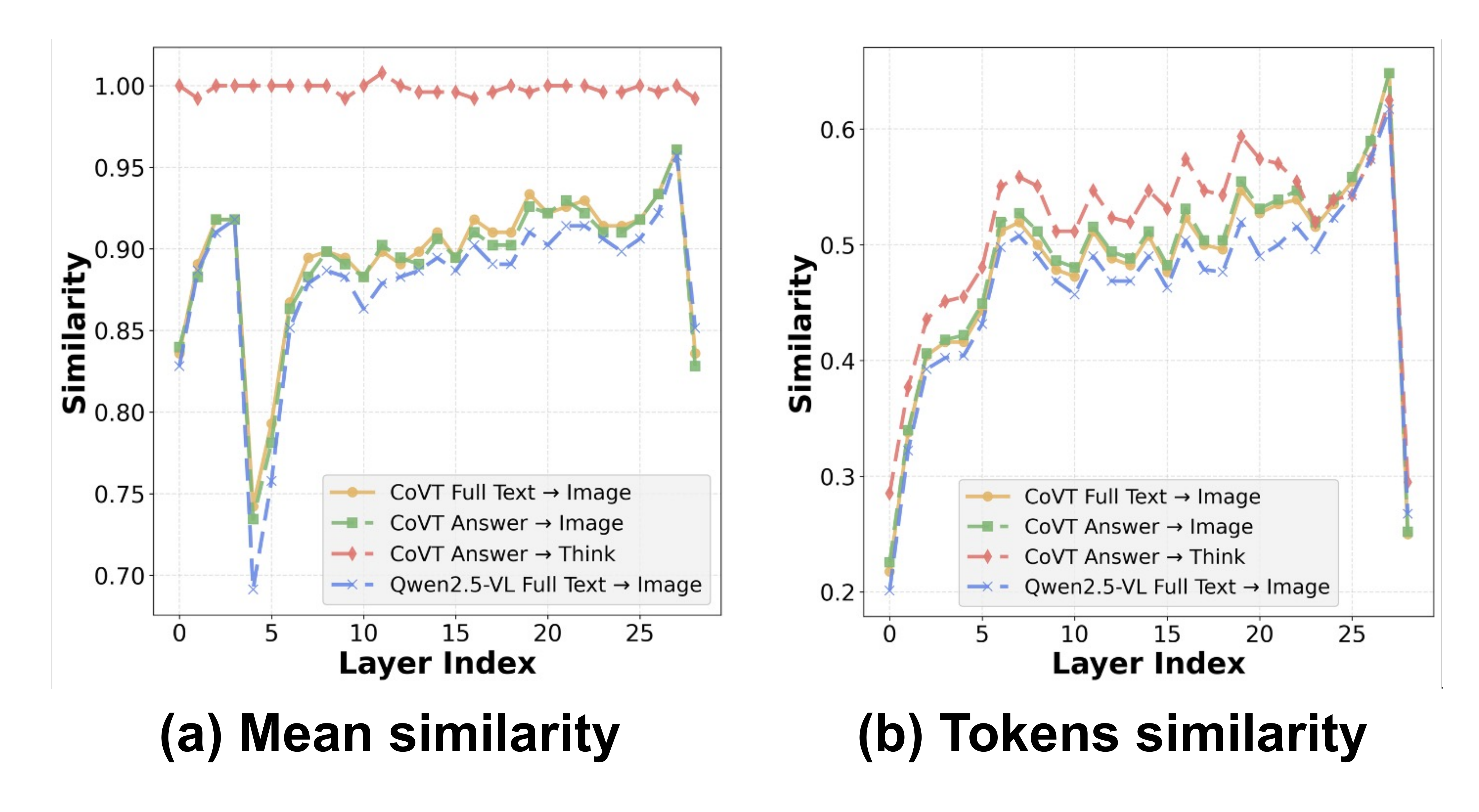}
\vspace{-0.5cm}
\captionof{figure}{\ours exhibits higher similarity between text tokens and image features.}
\label{fig:analysis-curve}
\end{minipage}
\hfill
\begin{minipage}{0.48\textwidth}
\centering
\includegraphics[width=\linewidth]{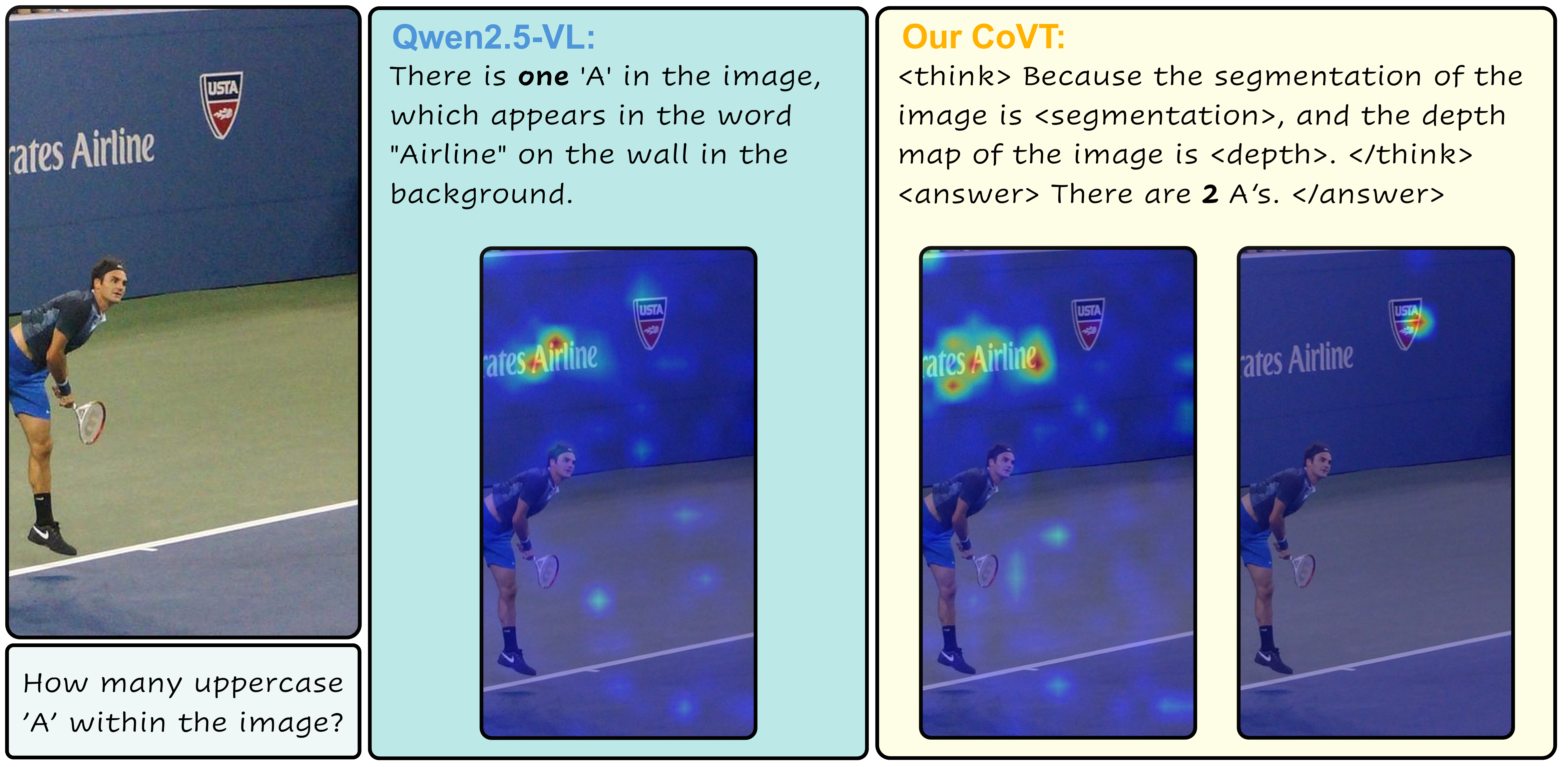}
\vspace{-0.5cm}
\captionof{figure}{In \ours, latent tokens across layers consistently attend to relevant regions correctly.}
\label{fig:analysis-attnmap}
\end{minipage}

\vspace{-0.4cm}
\end{strip}

\subsection{\ours Better Aligns with Visual Evidence}

We analyze \ours through (i) representation similarity between language tokens and visual features, and (ii) cross-attention patterns over the image.
In Fig.~\ref{fig:analysis-curve}, we compute the layer-wise cosine similarity between image features and different groups of output tokens from \ours, and compare against the baseline.
Concretely, we measure two kinds of similarity: (a) \textit{mean similarity}, the mean-pooled token similarity, and (b) \textit{tokens similarity}, the average of the tokens similarity, between (i) full output text and image features, (ii) answer tokens and image features, and (iii) answer tokens and intermediate ``thinking'' tokens.\footnote{All similarities are computed on the hidden states of the corresponding layer.}

Across intermediate layers, \ours shows consistently higher similarity between answer tokens and image features than the baseline, indicating stronger visual grounding throughout the forward pass. Moreover, \ours also has high similarity between answer tokens and thinking tokens, suggesting that its intermediate reasoning states are more predictive of the final answer and better aligned with the visual input.

We further inspect cross-attention from answer tokens to image tokens (Fig.~\ref{fig:analysis-attnmap}). Across layers, the baseline attends to only part of the relevant regions, whereas \ours more precisely and consistently covers the task-relevant regions, aligning with the baseline's failure and \ours's correct prediction.

\vspace{-0.1cm}
\section{Conclusions}

In this paper, we introduced \textbf{\ours}, the chain of continuous visual thoughts that enables vision–language models to reason beyond discrete linguistic space by leveraging compact, dense visual representations. 
\ours consistently improves visual-centric reasoning across diverse perception benchmarks and reveals that different types of visual tokens contribute to complementary aspects of multimodal understanding. 
These findings suggest that \ours can serve as a general framework for integrating fine-grained perceptual reasoning into broader multimodal systems.

\section{Acknowledgment}
We sincerely thank Himanshu Dubey and Sowmay Jain for API credit support from Anannas AI.
We greatly thank Baifeng Shi, Ji Xie, Shaofeng Yin for their insightful discussions and valuable feedback on our paper. We especially
thank Mahtab Bigverdi for her assistance in reproducing the baseline results.

{
    \small
    \bibliographystyle{ieeenat_fullname}
    \bibliography{main}
}

\appendix

\etocdepthtag.toc{mtappendix}
\etocsettagdepth{mtchapter}{none}    
\etocsettagdepth{mtappendix}{subsection} 

\clearpage
\maketitlesupplementary

\begingroup
  \parskip0pt
  \etocsettocstyle{\section*{Table of Contents}}{} 
  \tableofcontents
\endgroup


\section{Additional Details of \ours}
\label{sec:model-detail}

In this section, we provide a comprehensive description of \ours architecture. We first detail the design and functionality of the projection layer in (Sec.~\ref{sec:suppl-project}). Then we provide more specific alignment architecture about segmentation token, depth token, and edge token from (Sec.\ref{sec:suppl-sam}) to (Sec.\ref{sec:suppl-edge}). Finally, we provide the detailed composition of the dataset used in \ours in (Sec.~\ref{sec:suppl-dataset}).

\begin{figure}[h]
    \centering
    \includegraphics[width=1.0\linewidth]{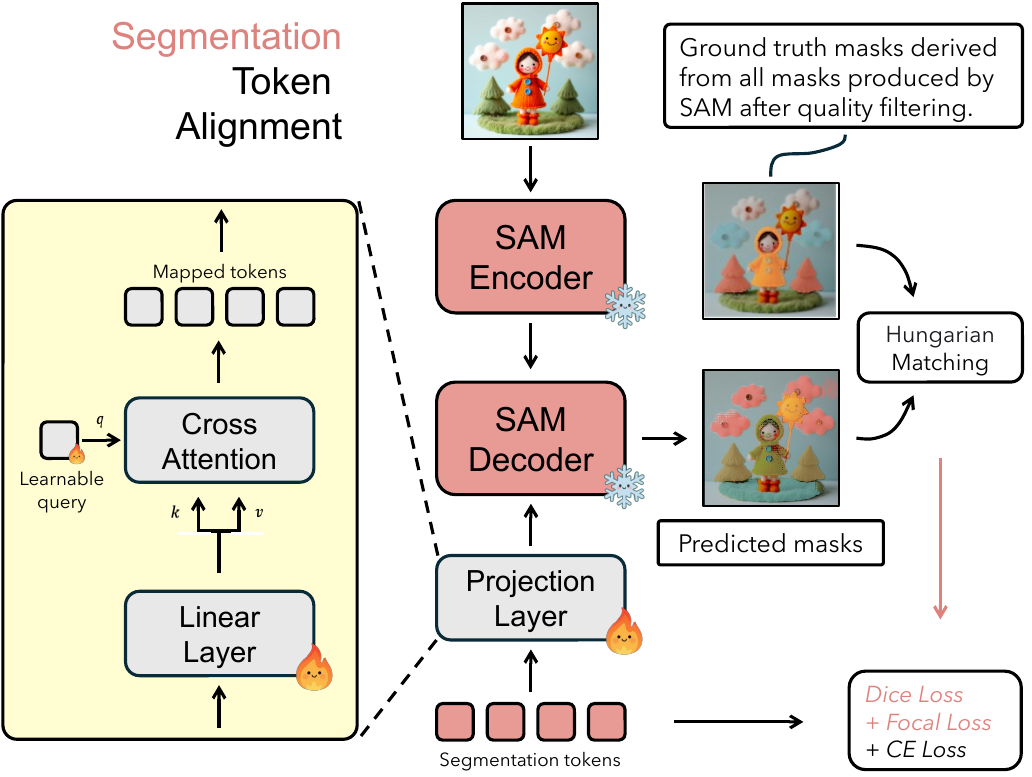}
    \caption{Detailed frameworks for the projection layer and segmentation token alignment.}
    \label{fig:suppl-project_sam}
\end{figure}

\subsection{Projection Layer}
\label{sec:suppl-project}

As shown in Fig.~\ref{fig:suppl-project_sam}, the details of the projection layer are illustrated in the yellow block. 
It contains a single linear layer that projects the VLM latent space into the decoder's prompt space (or the encoder's feature space while aligning DINO tokens), formulated as:
\begin{equation}
    \mathbf{z}_m = W \mathbf{z} + \mathbf{b},
\end{equation}
where $\mathbf{z}$ denotes the VLM latent feature and $\mathbf{z}_m$ is the mapped prompt-space feature after the linear layer.
We then introduce a learnable query $q$, while the mapped feature serves as both the key $k$ and value $v$ in the cross-attention layer, defined as:
\begin{equation}
    \mathbf{z}_p = \text{Attn}(q, k, v) = \text{softmax}\left( \frac{q k^\top}{\sqrt{d_k}} \right) v ,
\end{equation}
where $\mathbf{z}_p$ is the projected tokens, functioning as the prompts for the subsequent visual model decoding.

\begin{figure}[t]
    \centering
    \includegraphics[width=0.85\linewidth]{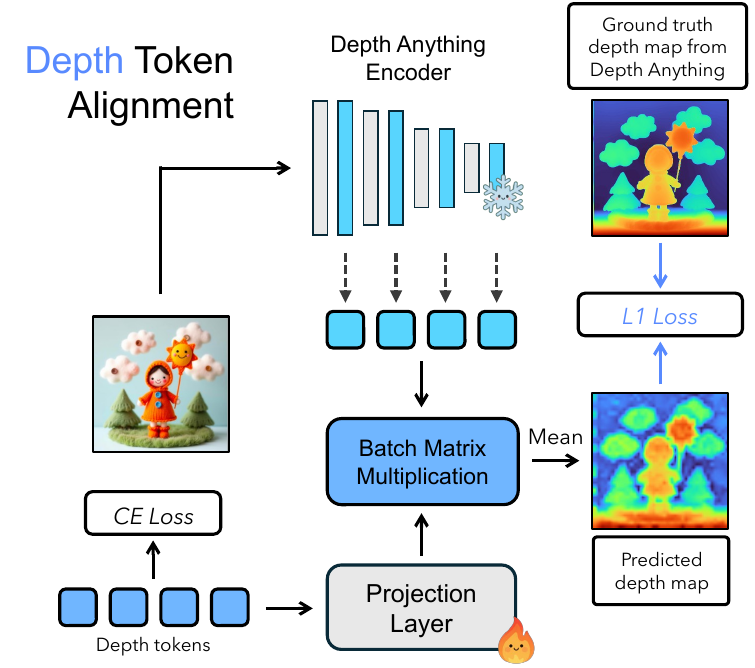}
    \caption{Detailed framework for the depth token alignment.}
    \label{fig:suppl-project_depth}
\end{figure}

\begin{figure}[t]
    \centering
    \includegraphics[width=1.0\linewidth]{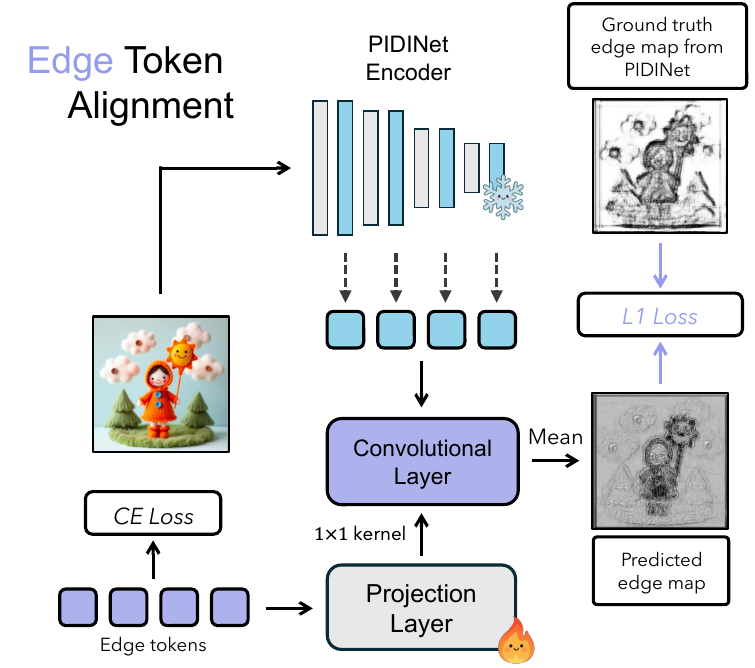}
    \caption{Detailed framework for edge token alignment.}
    \label{fig:suppl-project_edge}
\end{figure}

\subsection{Segmentation \ours Token Alignment}
\label{sec:suppl-sam}

Our model first predicts eight Segmentation tokens $\{T_i^{\text{seg}}\}_{i=0}^{7}$. Shown in Fig.~\ref{fig:suppl-project_sam}, each token is then projected into the SAM decoder's prompt space through the projection layer,
\begin{equation}
    T_i^{\text{sam}} = \text{proj}(T_i^{\text{seg}}) ,
\end{equation}
and the projected token serves as an individual prompt for mask decoding. Given the projected prompt $T_i^{\text{sam}}$ and the dense embedding $f$ from the SAM encoder, the SAM decoder produces one mask per token:
\begin{equation}
    \hat{M}_i = \text{Decoder}\!\left(T_i^{\text{sam}}, f\right),
    \qquad \hat{M}_i \in [0,1]^{H \times W}.
\end{equation}

To construct reliable supervision, we generate all masks from SAM on the input image and apply a quality filter based on \textit{mask area} and \textit{stability score}. From these, we retain eight high-quality masks,
\begin{equation}
    \mathcal{G} = \{ M_j \}_{j=0}^{7},
\end{equation}
which serve as ground truths.
We employ the Hungarian algorithm to match each predicted mask with one SAM mask. Unlike using similarity-based costs, we define the matching cost directly using the segmentation losses. For each pair $(\hat{M}_i, M_j)$, the Dice loss and Focal loss are
\begin{align}
    \mathcal{L}_{\text{dice}}\!(\hat{M}_i, M_j)
        &= 1 - \frac{2 \sum \hat{M}_i M_j}{
            \sum \hat{M}_i + \sum M_j}, \\
    \mathcal{L}_{\text{focal}}\!(\hat{M}_i, M_j)
        &= - (1 - \hat{M}_i)^{\gamma_F} M_j \log \hat{M}_i ,
\end{align}
where $\gamma_F$ is set to $2$. Therefore, the matching cost becomes
\begin{equation}
    C_{i,j}
        = \mathcal{L}_{\text{dice}}\!(\hat{M}_i, M_j)
        + \alpha \cdot \mathcal{L}_{\text{focal}}\!(\hat{M}_i, M_j),
\end{equation}
where the $\alpha$ is set to 1 in our experiments. The optimal assignment is then obtained via
\begin{equation}
    \sigma^{*} = \arg\min_{\sigma}
        \sum_{i=0}^{7} C_{i,\,\sigma(i)} .
\end{equation}
After obtaining the matching, the final mask loss is computed using the same Dice and Focal losses:
\begin{equation}
    \mathcal{L}_{\text{mask}}
        = \sum_{i=0}^{7}
            \left[
                \mathcal{L}_{\text{dice}}\!
                    \left(\hat{M}_i, M_{\sigma^{*}(i)}\right)
                +
                \alpha \cdot \mathcal{L}_{\text{focal}}\!
                    \left(\hat{M}_i, M_{\sigma^{*}(i)}\right)
            \right].
\end{equation}


\begin{table}[t]
\centering
\fontsize{8.5pt}{10pt}\selectfont
\begin{tabular}{lccc}
\toprule
 & \multicolumn{3}{c}{Qwen2.5-VL-7B} \\
\cmidrule(lr){2-4}
 & 1 type & 3 types & 4 types \\
\midrule
\textbf{Optimization} & & & \\
Optimizer & \multicolumn{3}{c}{AdamW} \\
Learning rate & \multicolumn{3}{c}{5e-5} \\
Projection layer lr & \multicolumn{3}{c}{1e-5} \\
lr schedule & \multicolumn{3}{c}{cosine} \\
$\beta$ & \multicolumn{3}{c}{(0.9, 0.999)} \\
Weight decay & \multicolumn{3}{c}{0.1} \\
Warmup ratio & \multicolumn{3}{c}{0.05} \\
First stage steps & 4K & 6K & 8K \\
Second stage steps & \multicolumn{3}{c}{3K} \\
Third stage steps & \multicolumn{3}{c}{3K} \\
Forth stage steps & 4K & 5K & 6K \\
Per-GPU batch size & \multicolumn{3}{c}{4} \\
$\gamma$ & \multicolumn{3}{c}{1.0} \\
\midrule
\textbf{LoRA settings} & & & \\
LoRA rank & \multicolumn{3}{c}{16} \\
LoRA alpha & \multicolumn{3}{c}{32} \\
\midrule
\textbf{Visual models} & & & \\
SAM Encoder & \multicolumn{3}{c}{ViT-H} \\
Depth Anything v2 Encoder & \multicolumn{3}{c}{ViT-L} \\
PIDINet Encoder & \multicolumn{3}{c}{Table5-Baseline} \\
DINO v2 & \multicolumn{3}{c}{ViT-L} \\
\bottomrule
\end{tabular}
\caption{Fine-tuning hyperparameter setup.}
\label{tab:suppl-exp_setup}
\end{table}

\subsection{Depth \ours Token Alignment}
\label{sec:suppl-depth}

As shown in Fig.~\ref{fig:suppl-project_depth}, our model predicts four Depth tokens $\{T_i^{\text{depth}}\}_{i=0}^{3}$, each of which is first projected into the DepthAnything decoder’s prompt space through a linear projection layer:
\begin{equation}
    T_i^{\text{depth-s}} = W_d T_i^{\text{depth}} + b_d .
\end{equation}
DepthAnything v2 provides four dense intermediate-layer features 
\[
    \{F_i^{\text{depth}}\}_{i=0}^{3},
\]
where $F_3^{\text{depth}}$ is the final-layer feature.  
Each projected depth token interacts with its corresponding feature map through batch matrix multiplication (BMM) to produce one depth map. This process is formulated as:
\begin{equation}
    \hat{D}_i = 
    \text{softmax}\!\left(
        T_i^{\text{depth-s}} \cdot {F_i^{\text{depth}}}^{\!\top}
    \right),
    \quad i = 0,\dots,3,
\end{equation}
where $\hat{D}_i$ denotes the $i$th reconstructed depth map.
Then the four reconstructed depth maps are averaged to produce the final prediction:
\begin{equation}
    \hat{D} = \frac{1}{4}\sum_{i=0}^{3} \hat{D}_i .
\end{equation}
For supervision, we use the depth map predicted by DepthAnything v2 as the ground truth, denoted as $D^{\text{gt}}$.  
Depth tokens are aligned through an L1 reconstruction loss between the final depth prediction and the ground truth:
\begin{equation}
    \mathcal{L}_{\text{depth}}
        = \left\| \hat{D} - D^{\text{gt}} \right\|_{1}.
\end{equation}

\begin{figure}[t]
    \centering
    \includegraphics[width=1.0\linewidth]{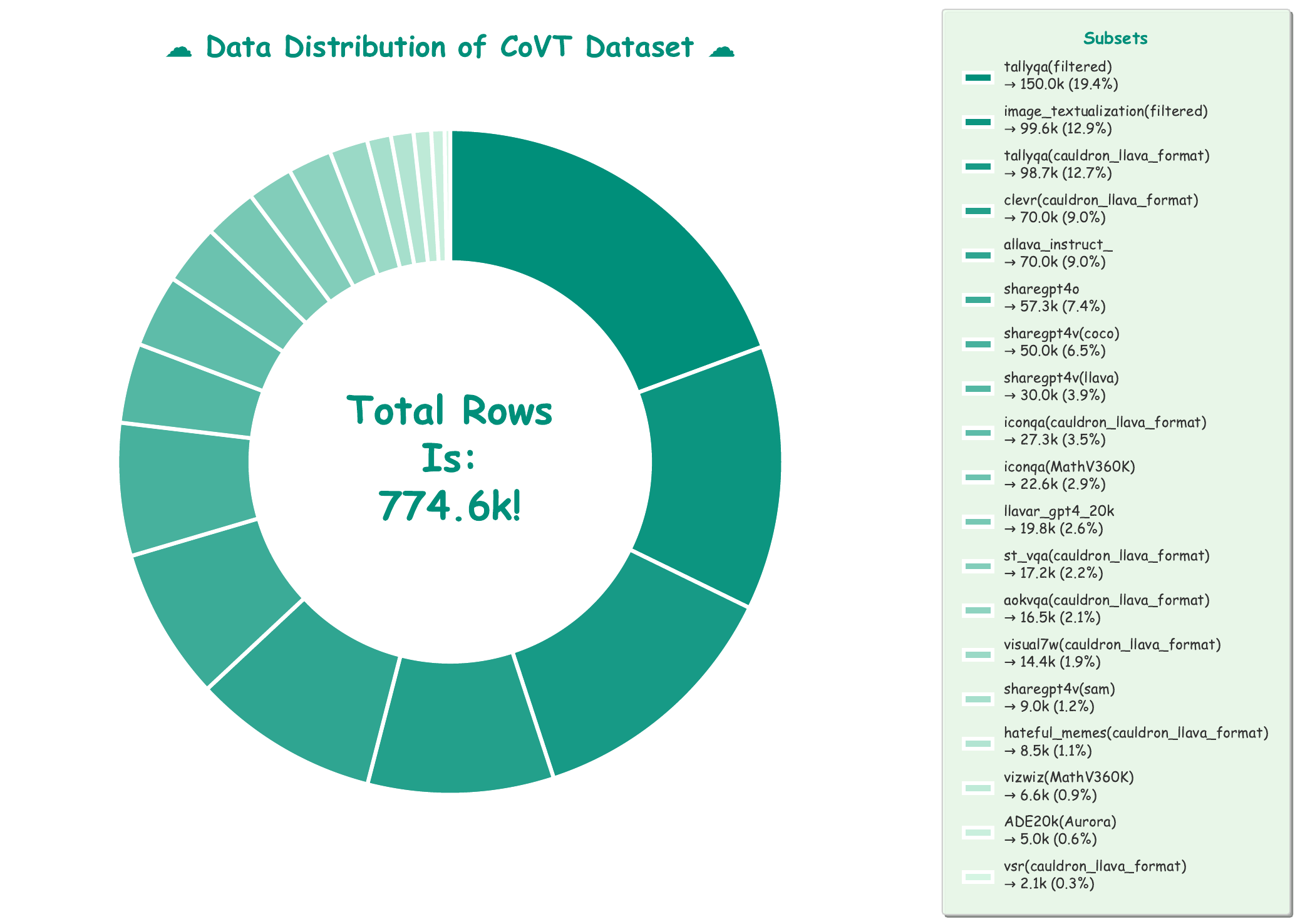}
    \captionsetup{width=1.0\linewidth}
    \caption{\ours dataset utilizes some subsets of LLaVA-OneVision, and merges the filtered TallyQA dataset and ADE20K-Depth from Aurora.}
    \label{fig:suppl-data}
\end{figure}

\subsection{Edge \ours Token Alignment}
\label{sec:suppl-edge}

Similarly, as shown in Fig.~\ref{fig:suppl-project_edge}, we first project the four predicted Edge tokens into the PIDINet prompt space, obtaining 
$\{T_i^{\text{edge}}\}_{i=0}^{3}$. Each projected token is then used as a $1 \times 1$ convolutional kernel and applied to the dense intermediate features extracted from the PIDINet encoder. Let $\{F_i^{\text{edge}}\}_{i=0}^{3}$ denote the four intermediate feature maps.  
For each level, the reconstructed edge map is obtained by
\begin{equation}
    \hat{E}_i = T_i^{\text{edge}} * F_i^{\text{edge}}, 
    \quad \hat{E}_i \in \mathbb{R}^{H \times W},
\end{equation}
where ``\(*\)'' denotes a $1 \times 1$ convolution operation. The four reconstructed edge maps are then aggregated by averaging, followed by a sigmoid normalization:
\begin{equation}
    \hat{E}
        = \sigma\!\left( \frac{1}{4} \sum_{i=0}^{3} \hat{E}_i \right).
\end{equation}
For supervision, we use the edge map predicted directly by PIDINet,
denoted as $E^{\text{gt}}$.  The alignment between the predicted and ground-truth edges is enforced using the L1 loss:
\begin{equation}
    \mathcal{L}_{\text{edge}}
        = \left\| \hat{E} - E^{\text{gt}} \right\|_{1}.
\end{equation}


\subsection{\ours Dataset Composition}
\label{sec:suppl-dataset}

In order to fully leverage the value of \ours, we select the vision-centric subsets from the LLaVA-OneVision dataset, including: \begin{itemize}
\item \textit{IconQA(MathV360K)}
\item \textit{VizWiz(MathV360K)}
\item \textit{allava\_instruct\_}
\item \textit{aokvqa(cauldron\_llava\_format)}
\item \textit{clevr(cauldron\_llava\_format)}
\item \textit{hateful\_memes(cauldron\_llava\_format)}
\item \textit{iconqa(cauldron\_llava\_format)}
\item \textit{image\_textualization(filtered)}
\item \textit{llavar\_gpt4\_20k}
\item \textit{sharegpt4o}
\item \textit{sharegpt4v(coco)}
\item \textit{sharegpt4v(llava)}
\item \textit{sharegpt4v(sam)}
\item \textit{st\_vqa(cauldron\_llava\_format)}
\item \textit{tallyqa(cauldron\_llava\_format)}
\item \textit{visual7w(cauldron\_llava\_format)}
\item \textit{vsr(cauldron\_llava\_format)}
\end{itemize}

In addition, we re-filtered the TallyQA dataset. Since TallyQA is a counting dataset but contains many samples with the answer 0, we reduce the proportion of zero-count samples and construct a 150k-sample subset. Moreover, following the methodology used in the Aurora paper, we generate 5k samples related to relative depth from the ADE20K dataset using the same procedure. We integrate these three components to form the complete \ours dataset.


\section{Additional Experiments}
\label{sec:suppl-exp}

In this section, we first describe the experimental settings used throughout our study in (Sec.~\ref{sec:suppl-setting}). 
We then present ablation studies on the first two training stages in (Sec.~\ref{sec:suppl-stage}).
Subsequently, we investigate the impact of varying the number of \ours tokens in (Sec.~\ref{sec:suppl-number}). 
Finally, we provide additional output examples in (Sec.~\ref{sec:suppl-result}).


\subsection{More settings}
\label{sec:suppl-setting}

In Tab.~\ref{tab:suppl-exp_setup}, we present the complete hyperparameter configurations to ensure full reproducibility of our experiments. 
In this table, \emph{1 type} denotes that the model is aligned with a single supervision signal, chosen from segmentation, depth, or DINO tokens. 
\emph{3 types} corresponds to jointly aligning the model with segmentation, depth, and DINO tokens. 
\emph{4 types} further incorporates edge tokens, thereby enabling simultaneous alignment across segmentation, depth, DINO, and edge tokens. 
For hyperparameters that remain consistent across all three experimental settings, we consolidate the corresponding columns and report them using a single centered entry for clarity and conciseness.

\begin{table}[t]
\centering
\fontsize{8.1pt}{10pt}\selectfont
\setlength{\tabcolsep}{2.0pt}
\renewcommand\arraystretch{0.95}
\newcommand{\pos}[1]{\textcolor{green!60!black}{#1}}
\newcommand{\negval}[1]{\textcolor{red!70!black}{#1}}
\begin{tabular}{lcccccccc}
\toprule
Quantity & CVBench & BLINK & RW-QA & MM*-P & MMVP & V* & $\text{HR}_{4K}$ \\
\midrule
Stage 3 \& 4     & 78.2 & 53.8 & 70.0 & 68.3 & \textbf{60.7} & \textbf{78.0} & 71.2 \\
\rowcolor{yellow!12}
Ours     & \textbf{80.0} & \textbf{56.0} & \textbf{71.6} & \textbf{69.2} & 58.7 & \textbf{78.0} & \textbf{72.9} \\
\bottomrule
\end{tabular}
\vspace{-0.2cm}
\caption{The first two stages in \ours 4-stage training strategy enhance the stability of the improvement.}
\vspace{-0.5cm}
\label{tab:suppl-ablation_stage}
\end{table}

\subsection{Training Stage Impact Ablation}
\label{sec:suppl-stage}

To elucidate the pivotal contribution of the first two stages in the four-stage training strategy of \ours, we perform an ablation study comparing the full model with a variant trained solely on Stages 3 and 4, as reported in Tab.~\ref{tab:suppl-ablation_stage}. 
The model trained across all four stages exhibits consistent and robust improvements over all evaluated benchmarks. 
In contrast, when restricting training to only the last two stages, \ours experiences notable degradation on the \textit{BLINK} benchmark and yields only marginal gains on \textit{RealWorld-QA} and \textit{MMStar-Perception}. 
These results underscore the critical role of the early-stage training signals and highlight their importance in shaping the model's downstream performance.


\subsection{Token Numbers Ablation}
\label{sec:suppl-number}

\begin{figure}[t]
    \centering
    \includegraphics[width=1.0\linewidth]{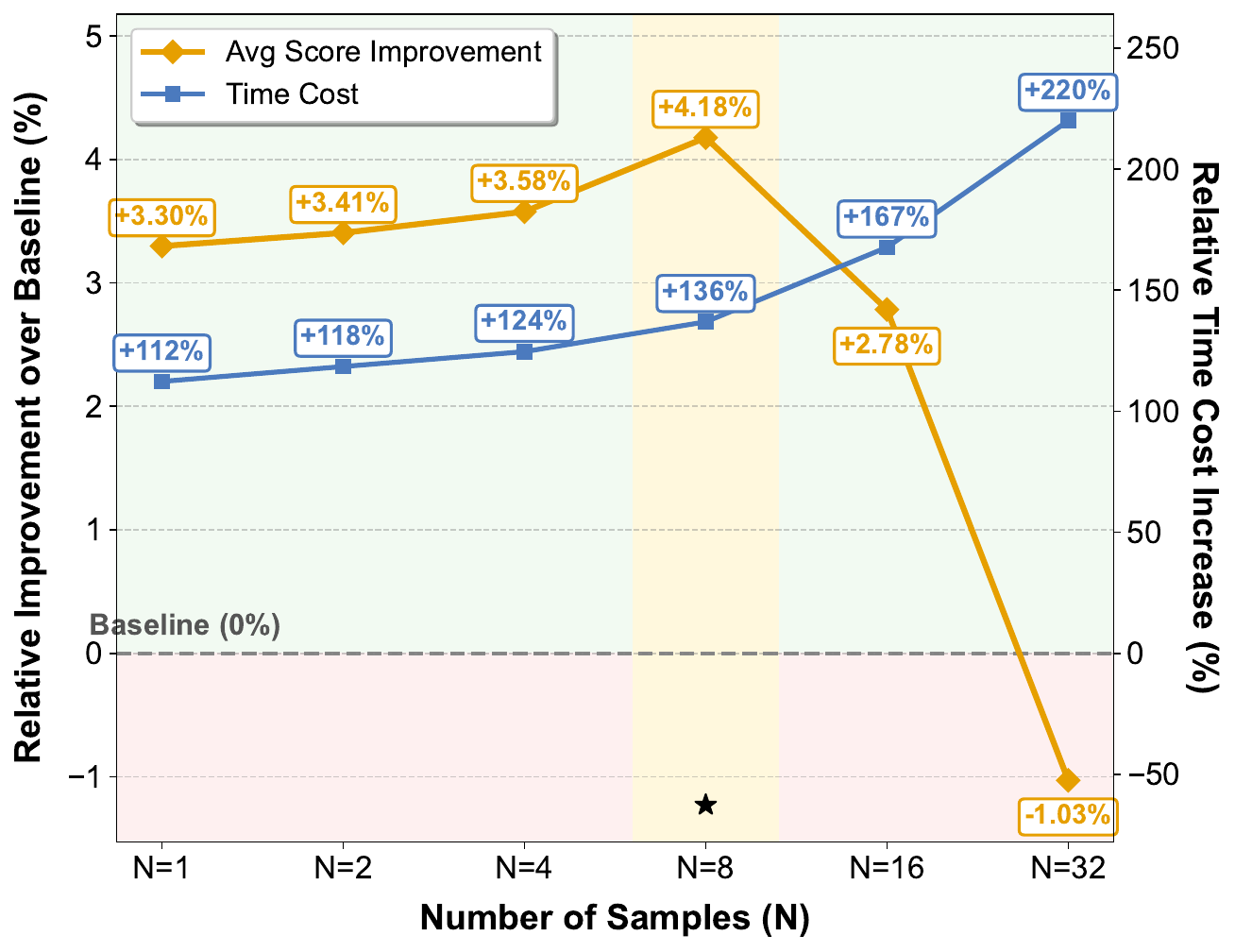}
    \caption{Equipped with 4 depth tokens and 4 DINO tokens, the model achieves its best performance when using 8 segmentation tokens in \ours. Allocating more \ours tokens leads to a slight diminishing performance and increased computational cost.}
    \label{fig:suppl-ablat_num}
    \vspace{-0.6cm}
\end{figure}

To determine the optimal number of segmentation tokens, we conduct a detailed ablation while fixing both depth and DINO tokens at 4. As shown in Fig.~\ref{fig:suppl-ablat_num}, increasing the number of segmentation tokens in \ours from 1 to 32 yields an initial performance gain followed by a gradual decline, whereas the computational overhead rises steadily. Notably, the overall time cost of \ours is considerably higher than that of the baseline. Only a minor fraction of this overhead originates from the additional \ours tokens; a more substantial portion stems from \ours producing richer and more fine-grained responses for tasks such as image captioning, which naturally leads to longer output sequences (examples provided in Sec.~\ref{sec:suppl-result}). Under our experimental conditions, eight segmentation tokens—combined with four depth tokens and four DINO tokens—offer the most favorable balance between performance and efficiency.

\subsection{More Results}
\label{sec:suppl-result}

\begin{figure*}
    \centering
    \includegraphics[width=0.58\linewidth]{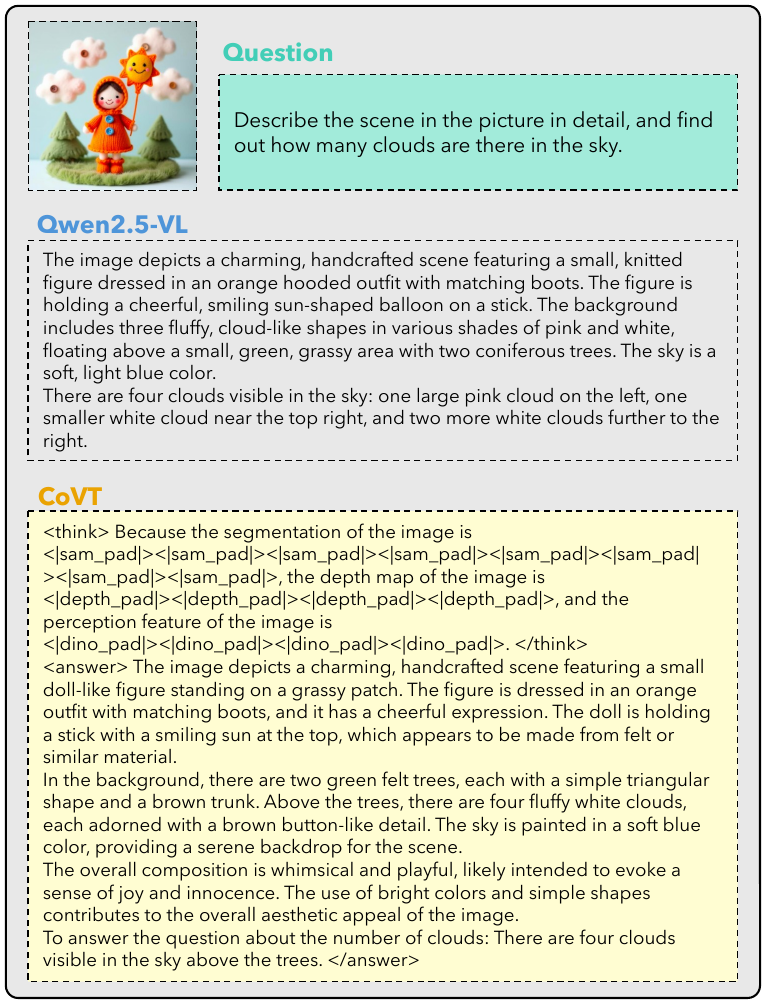}
    \caption{Example of \ours compared to the baseline Qwen2.5-VL-7B.}
    \label{fig:suppl-example1}
\end{figure*}

\begin{figure*}
    \centering
    \includegraphics[width=0.45\linewidth]{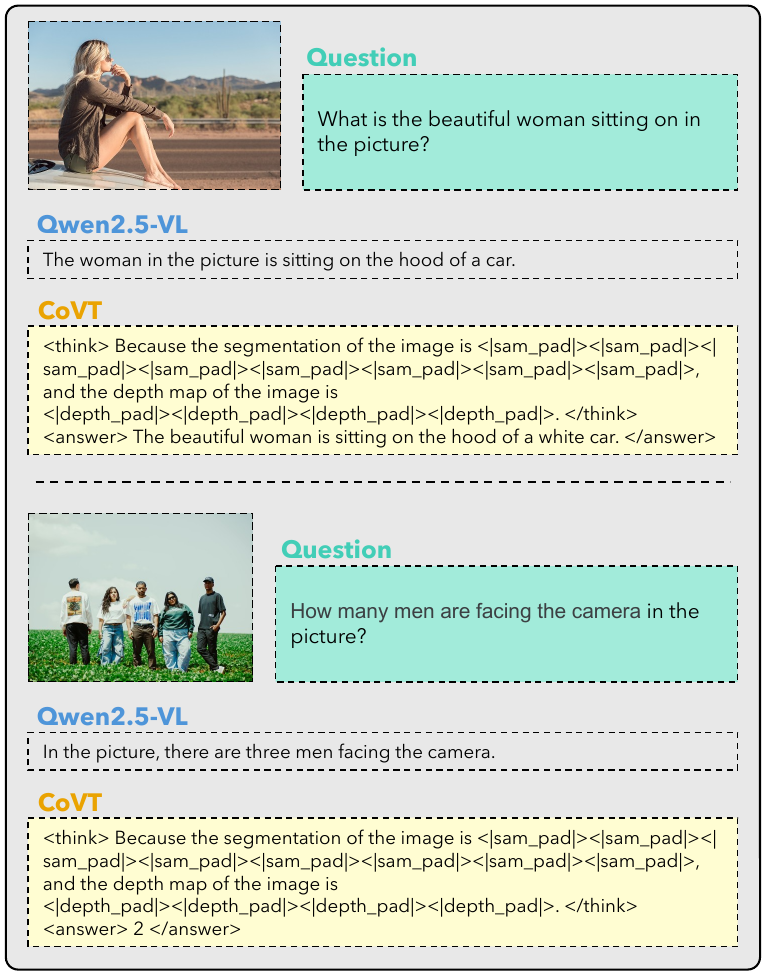}
    \caption{Examples of \ours compared to the baseline Qwen2.5-VL-7B.}
    \label{fig:suppl-example2}
\end{figure*}

\begin{figure*}
    \centering
    \includegraphics[width=0.45\linewidth]{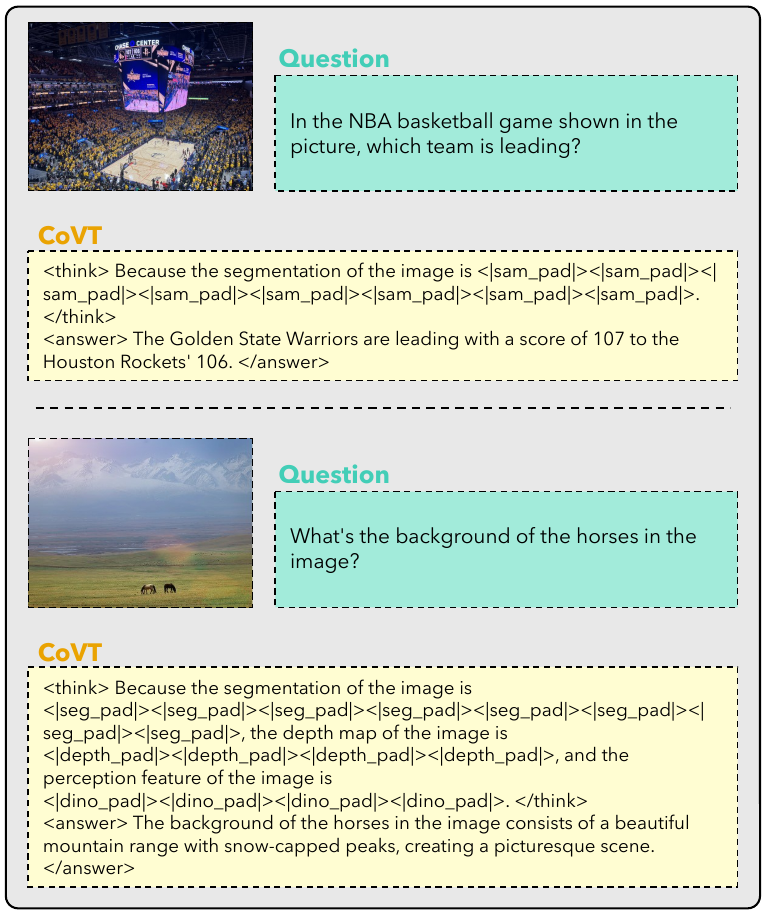}
    \caption{More examples of \ours.}
    \label{fig:suppl-example3}
\end{figure*}

\begin{figure*}
    \centering
    \includegraphics[width=0.495\linewidth]{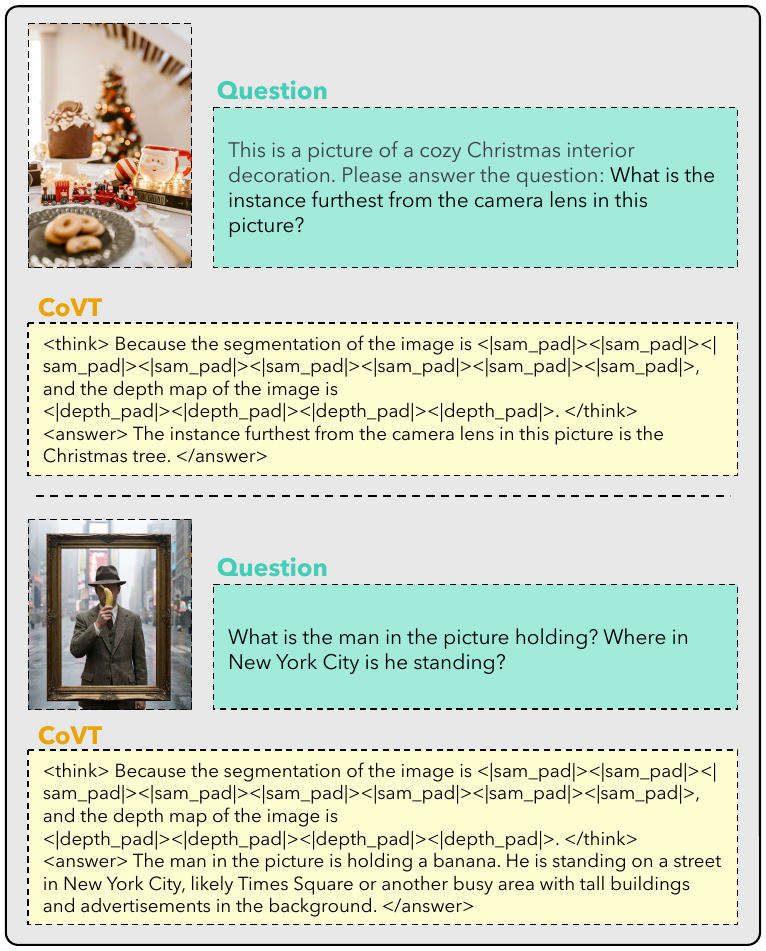}
    \caption{More examples of \ours.}
    \label{fig:suppl-example4}
\end{figure*}

\begin{figure*}
    \centering
    \includegraphics[width=0.5\linewidth]{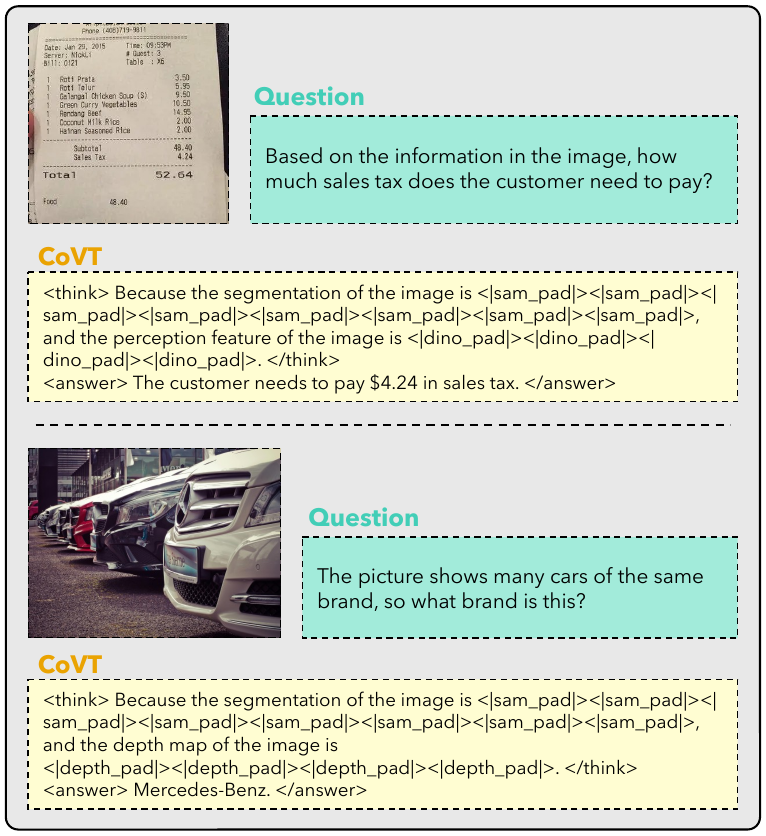}
    \caption{More examples of \ours.}
    \label{fig:suppl-example5}
\end{figure*}

We provide additional VQA examples in Fig.~\ref{fig:suppl-example1} through Fig.~\ref{fig:suppl-example5}, including detailed image captioning, counting, instance identification, depth-aware questions, real-world OCR, and so on. In Fig.~\ref{fig:suppl-example1}, we compare \ours with the baseline and show that \ours offers more fine-grained captioning ability. In Fig.~\ref{fig:suppl-example2}, \ours improves instance identification (\textit{e.g.}, describing an object as the white car hood) and better counting performance. In the subsequent figures from Fig.~\ref{fig:suppl-example3} to Fig.~\ref{fig:suppl-example5}, we demonstrate that our model maintains stable performance across various vision-centric tasks and is also capable of tackling text-centric tasks such as real-world OCR. 
Specifically, in Fig.~\ref{fig:suppl-example3}, \ours demonstrates its ability to correctly identify the NBA teams and their scores in the first example, and to recognize visually ambiguous backgrounds in the second example. In Fig.~\ref{fig:suppl-example4}, the first example shows that \ours can identify the farthest object and classify it correctly, while the second example illustrates that \ours can handle common-sense VQA tasks (e.g., identifying that the tall buildings are from Times Square). The last figure, Fig.~\ref{fig:suppl-example5}, shows that \ours maintains stable performance on OCR tasks. In the first example, the model accurately detects the text “Sales Tax 4.24”, and in the second example, it identifies the partially visible text “Mer” in the background, along with several car logos on the side.

\section{Limitations and Future Work}
Despite the strong performance of \ours, several limitations remain. 
First, \textit{{we have not exhaustively explored the design space of visual models or token combinations.}} 
Our current setup focuses on representative perception axes such as segmentation, depth, edges, and DINO features, but alternative or hybrid visual experts may yield more expressive or complementary tokens. 
Systematic exploration of this space could further enhance reasoning performance. 
Second, \textit{{our current framework does not yet realize fully interleaved multimodal reasoning.}} 
\ours generates continuous visual thoughts but does not integrate them with free-form textual reasoning in an interleaved sequence. 
Developing such a unified chain that seamlessly blends textual and visual thoughts represents a promising direction for future work.

We believe that \ours provides a foundation for the next generation of multimodal reasoning systems—those capable of thinking fluidly across both language and vision in a self-contained, interpretable manner.

\end{document}